%% file: TCAD_ResistModelAL.tex
\documentclass[journal,9pt]{IEEEtran}
\input{./ieee-normal}

\graphicspath{{../figs/}{./}}

\definecolor{mygreen}{HTML}{4BC3AA}
\definecolor{myblue}{HTML}{143EF7}
\definecolor{mygrey}{HTML}{73FDFF}
\definecolor{aspar}{RGB}{178,127,0}
\definecolor{sky}{RGB}{102,204,255}
\definecolor{flora}{RGB}{102,255,102}
\definecolor{viola}{RGB}{153,0,153}

\newcommand{\DefMacro}[2]{\expandafter\newcommand\csname rmk-#1\endcsname{#2}}
\newcommand{\UseMacro}[1]{\csname rmk-#1\endcsname}

\usepackage[nomessages]{fp}

\def\DeltaScale#1#2#3{%
    \FPeval\result{clip(((#1)-(#2))*(#3))}%
\relax{\result}%
}
\def\RatioScale#1#2#3{%
    \FPeval\result{clip(((#1)/(#2))*(#3))}%
\relax{\result}%
}

\begin{document}

\title{
    Data Efficient Lithography Modeling with Transfer Learning and Active Data Selection
}

\author{
Yibo Lin~\IEEEmembership{Student Member,~IEEE},
Meng Li~\IEEEmembership{Student Member,~IEEE}, 
Yuki Watanabe,
Taiki Kimura, 
Tetsuaki Matsunawa, 
Shigeki Nojima, 
David Z.~Pan~\IEEEmembership{Fellow,~IEEE}\\
\thanks{Y.~Lin, M.~Li and D.~Z.~Pan are with The Department of Electrical and Computer Engineering, The University of Texas at Austin, TX, USA.}
\thanks{Y.~Watanabe, T.~Kimura, T.~Matsunawa, and S.~Nojima are with Toshiba Memory Corporation, Yokohama, Japan.}
}

\maketitle
\thispagestyle{empty} 

\input{data10AverageMacros}
\input{abstract}

\input{intro}

\input{prelim}

\input{algo}
\input{result}
\input{conclu}

{
\bibliographystyle{IEEEtran}
\bibliography{./Top,./Software,./DFM,./Cell,./ALG,./PD,./addition}
}

\input{appdx}

\end{document}

%% file: ieee-normal.tex
\usepackage{algpseudocode}                                 
\usepackage{algorithm}
\usepackage{graphicx}                                      
\usepackage{amsmath}
\usepackage{amssymb}
\usepackage{amsfonts}
\usepackage{booktabs}
\usepackage{multirow}
\usepackage[subrefformat=parens,farskip=0pt,justification=centering]{subfig}
\usepackage{color}
\usepackage{cite}                                          
\usepackage{comment}                                       
\usepackage{soul}                                          
\soulregister\cite7
\soulregister\ref7
\soulregister\pageref7
\usepackage{amsthm}
\usepackage{etoolbox}                                      
\usepackage{url}
\usepackage{nth}                                           
\usepackage{bm}                                            
\usepackage{courier}
\usepackage{balance}
\usepackage{threeparttable}
\usepackage[bookmarks=false]{hyperref}                     %
\hypersetup{
    colorlinks = true,
    citecolor  = blue,
    linkcolor  = blue,
    urlcolor   = blue,
}
\usepackage{filecontents}                                  
\usepackage{tikz}
\usetikzlibrary{positioning, fit}
\usepackage{pgfplots}
\usepackage{pgfplotstable}
\pgfplotsset{compat=newest}
\usepackage{caption} 
\definecolor{CUHKorange}{RGB}{244,106,18} 
\definecolor{CUHKblue}{RGB}{0,111,190}    
\definecolor{CUHKgreen}{RGB}{0,127,128}   
\definecolor{CUHKred}{RGB}{228,46,36}     
\definecolor{CUHKyellow}{RGB}{198,148,34} 
\definecolor{CUHKdark}{RGB}{114,44,114}   
\definecolor{CUHKmiddle}{RGB}{144,44,144} 
\definecolor{CUHKlight}{RGB}{167,44,167} 

\newtheorem{theorem}{Theorem}[section]
\newtheorem{lemma}[theorem]{Lemma}
\newtheorem{definition}[theorem]{Definition}

\newtheorem{problem}[theorem]{Problem}

\newcommand{\norm}[2][\defmu]{{\left\lVert {#2} \right\rVert}_{#1}}
\newcommand{\abs}[2][\defmu]{{\left\lvert {#2} \right\rvert}_{#1}}

\algrenewcommand\textproc{\texttt}

\makeatletter
\let\OldStatex\Statex
\renewcommand{\Statex}[1][3]{%
  \setlength\@tempdima{\algorithmicindent}%
  \OldStatex\hskip\dimexpr#1\@tempdima\relax
}
\makeatother

\RequirePackage[normalem]{ulem} 
\RequirePackage{color}\definecolor{RED}{rgb}{1,0,0}\definecolor{BLUE}{rgb}{0,0,1} 

%% file: data10AverageMacros.tex
\DefMacro{N10ToN7Final.ResNN.transferFromN10P50.CNN.0.01.imp}{52}
\DefMacro{N10ToN7Final.ResNN.transferFromN10P50.CNN.0.2.imp}{18}
\DefMacro{N10ToN7Final.ResNN.transferFromN10P50.CNN.transferFromN10P50.avg}{13}
\DefMacro{N7ToN7Final.ResNN.transferFromN7V20P50.CNN.0.01.imp}{65}
\DefMacro{N7ToN7Final.ResNN.transferFromN7V20P50.CNN.0.2.imp}{23}
\DefMacro{N7ToN7Final.ResNN.transferFromN7V20P50.CNN.transferFromN7V20P50.avg}{15}
\DefMacro{N7ToN7Final.ResNN.transferFix8FromN7V20P50.ResNN.transferFromN7V20P50.max}{28}
\DefMacro{N7ToN7Final.ResNN.transferFix8FromN7V20P50.ResNN.transferFromN7V20P50.min}{14}

\DefMacro{N7ToN7Final.CNN.N7.V23.0.01.thresholdPredictRelativeRMS}{4.44}
\DefMacro{N7ToN7Final.CNN.N7.V23.0.01.CDPredictRMS}{4.76}
\DefMacro{N7ToN7Final.CNN.N7.V23.0.05.thresholdPredictRelativeRMS}{2.78}
\DefMacro{N7ToN7Final.CNN.N7.V23.0.05.CDPredictRMS}{2.96}
\DefMacro{N7ToN7Final.CNN.N7.V23.0.1.thresholdPredictRelativeRMS}{1.92}
\DefMacro{N7ToN7Final.CNN.N7.V23.0.1.CDPredictRMS}{2.04}
\DefMacro{N7ToN7Final.CNN.N7.V23.0.15.thresholdPredictRelativeRMS}{1.72}
\DefMacro{N7ToN7Final.CNN.N7.V23.0.15.CDPredictRMS}{1.84}
\DefMacro{N7ToN7Final.CNN.N7.V23.0.2.thresholdPredictRelativeRMS}{1.60}
\DefMacro{N7ToN7Final.CNN.N7.V23.0.2.CDPredictRMS}{1.71}
\DefMacro{N7ToN7Final.CNN.N7.V23.0.3.thresholdPredictRelativeRMS}{1.44}
\DefMacro{N7ToN7Final.CNN.N7.V23.0.3.CDPredictRMS}{1.55}
\DefMacro{N7ToN7Final.CNN.N7.V23.0.4.thresholdPredictRelativeRMS}{1.35}
\DefMacro{N7ToN7Final.CNN.N7.V23.0.4.CDPredictRMS}{1.45}
\DefMacro{N7ToN7Final.CNN.N7.V23.0.5.thresholdPredictRelativeRMS}{1.24}
\DefMacro{N7ToN7Final.CNN.N7.V23.0.5.CDPredictRMS}{1.33}

\DefMacro{N10ToN7Final.CNN.transferFromN10P50.N7.V23.0.01.thresholdPredictRelativeRMS}{2.34}
\DefMacro{N10ToN7Final.CNN.transferFromN10P50.N7.V23.0.01.CDPredictRMS}{2.48}
\DefMacro{N10ToN7Final.CNN.transferFromN10P50.N7.V23.0.05.thresholdPredictRelativeRMS}{1.73}
\DefMacro{N10ToN7Final.CNN.transferFromN10P50.N7.V23.0.05.CDPredictRMS}{1.86}
\DefMacro{N10ToN7Final.CNN.transferFromN10P50.N7.V23.0.1.thresholdPredictRelativeRMS}{1.63}
\DefMacro{N10ToN7Final.CNN.transferFromN10P50.N7.V23.0.1.CDPredictRMS}{1.76}
\DefMacro{N10ToN7Final.CNN.transferFromN10P50.N7.V23.0.15.thresholdPredictRelativeRMS}{1.56}
\DefMacro{N10ToN7Final.CNN.transferFromN10P50.N7.V23.0.15.CDPredictRMS}{1.68}
\DefMacro{N10ToN7Final.CNN.transferFromN10P50.N7.V23.0.2.thresholdPredictRelativeRMS}{1.50}
\DefMacro{N10ToN7Final.CNN.transferFromN10P50.N7.V23.0.2.CDPredictRMS}{1.61}
\DefMacro{N10ToN7Final.CNN.transferFromN10P50.N7.V23.0.3.thresholdPredictRelativeRMS}{1.39}
\DefMacro{N10ToN7Final.CNN.transferFromN10P50.N7.V23.0.3.CDPredictRMS}{1.49}
\DefMacro{N10ToN7Final.CNN.transferFromN10P50.N7.V23.0.4.thresholdPredictRelativeRMS}{1.31}
\DefMacro{N10ToN7Final.CNN.transferFromN10P50.N7.V23.0.4.CDPredictRMS}{1.41}
\DefMacro{N10ToN7Final.CNN.transferFromN10P50.N7.V23.0.5.thresholdPredictRelativeRMS}{1.24}
\DefMacro{N10ToN7Final.CNN.transferFromN10P50.N7.V23.0.5.CDPredictRMS}{1.32}

\DefMacro{N10ToN7Final.ResNN.transferFromN10P50.N7.V23.0.01.thresholdPredictRelativeRMS}{2.29}
\DefMacro{N10ToN7Final.ResNN.transferFromN10P50.N7.V23.0.01.CDPredictRMS}{2.39}
\DefMacro{N10ToN7Final.ResNN.transferFromN10P50.N7.V23.0.05.thresholdPredictRelativeRMS}{1.60}
\DefMacro{N10ToN7Final.ResNN.transferFromN10P50.N7.V23.0.05.CDPredictRMS}{1.70}
\DefMacro{N10ToN7Final.ResNN.transferFromN10P50.N7.V23.0.1.thresholdPredictRelativeRMS}{1.47}
\DefMacro{N10ToN7Final.ResNN.transferFromN10P50.N7.V23.0.1.CDPredictRMS}{1.57}
\DefMacro{N10ToN7Final.ResNN.transferFromN10P50.N7.V23.0.15.thresholdPredictRelativeRMS}{1.39}
\DefMacro{N10ToN7Final.ResNN.transferFromN10P50.N7.V23.0.15.CDPredictRMS}{1.47}
\DefMacro{N10ToN7Final.ResNN.transferFromN10P50.N7.V23.0.2.thresholdPredictRelativeRMS}{1.31}
\DefMacro{N10ToN7Final.ResNN.transferFromN10P50.N7.V23.0.2.CDPredictRMS}{1.39}
\DefMacro{N10ToN7Final.ResNN.transferFromN10P50.N7.V23.0.3.thresholdPredictRelativeRMS}{1.19}
\DefMacro{N10ToN7Final.ResNN.transferFromN10P50.N7.V23.0.3.CDPredictRMS}{1.26}
\DefMacro{N10ToN7Final.ResNN.transferFromN10P50.N7.V23.0.4.thresholdPredictRelativeRMS}{1.09}
\DefMacro{N10ToN7Final.ResNN.transferFromN10P50.N7.V23.0.4.CDPredictRMS}{1.15}
\DefMacro{N10ToN7Final.ResNN.transferFromN10P50.N7.V23.0.5.thresholdPredictRelativeRMS}{0.98}
\DefMacro{N10ToN7Final.ResNN.transferFromN10P50.N7.V23.0.5.CDPredictRMS}{1.04}

\DefMacro{ActiveLearning.KMeans.ResNN.transferFromN10P50.N7.V23.0.01.thresholdPredictRelativeRMS}{1.95}
\DefMacro{ActiveLearning.KMeans.ResNN.transferFromN10P50.N7.V23.0.01.CDPredictRMS}{2.09}
\DefMacro{ActiveLearning.KMeans.ResNN.transferFromN10P50.N7.V23.0.05.thresholdPredictRelativeRMS}{1.58}
\DefMacro{ActiveLearning.KMeans.ResNN.transferFromN10P50.N7.V23.0.05.CDPredictRMS}{1.70}
\DefMacro{ActiveLearning.KMeans.ResNN.transferFromN10P50.N7.V23.0.1.thresholdPredictRelativeRMS}{1.36}
\DefMacro{ActiveLearning.KMeans.ResNN.transferFromN10P50.N7.V23.0.1.CDPredictRMS}{1.48}
\DefMacro{ActiveLearning.KMeans.ResNN.transferFromN10P50.N7.V23.0.15.thresholdPredictRelativeRMS}{1.23}
\DefMacro{ActiveLearning.KMeans.ResNN.transferFromN10P50.N7.V23.0.15.CDPredictRMS}{1.32}
\DefMacro{ActiveLearning.KMeans.ResNN.transferFromN10P50.N7.V23.0.2.thresholdPredictRelativeRMS}{1.16}
\DefMacro{ActiveLearning.KMeans.ResNN.transferFromN10P50.N7.V23.0.2.CDPredictRMS}{1.24}
\DefMacro{ActiveLearning.KMeans.ResNN.transferFromN10P50.N7.V23.0.3.thresholdPredictRelativeRMS}{1.05}
\DefMacro{ActiveLearning.KMeans.ResNN.transferFromN10P50.N7.V23.0.3.CDPredictRMS}{1.12}
\DefMacro{ActiveLearning.KMeans.ResNN.transferFromN10P50.N7.V23.0.4.thresholdPredictRelativeRMS}{1.00}
\DefMacro{ActiveLearning.KMeans.ResNN.transferFromN10P50.N7.V23.0.4.CDPredictRMS}{1.07}

\DefMacro{N7ToN7Final.CNN.transferFromN7V20P50.N7.V23.0.01.thresholdPredictRelativeRMS}{1.69}
\DefMacro{N7ToN7Final.CNN.transferFromN7V20P50.N7.V23.0.01.CDPredictRMS}{1.79}
\DefMacro{N7ToN7Final.CNN.transferFromN7V20P50.N7.V23.0.05.thresholdPredictRelativeRMS}{1.53}
\DefMacro{N7ToN7Final.CNN.transferFromN7V20P50.N7.V23.0.05.CDPredictRMS}{1.64}
\DefMacro{N7ToN7Final.CNN.transferFromN7V20P50.N7.V23.0.1.thresholdPredictRelativeRMS}{1.50}
\DefMacro{N7ToN7Final.CNN.transferFromN7V20P50.N7.V23.0.1.CDPredictRMS}{1.60}
\DefMacro{N7ToN7Final.CNN.transferFromN7V20P50.N7.V23.0.15.thresholdPredictRelativeRMS}{1.48}
\DefMacro{N7ToN7Final.CNN.transferFromN7V20P50.N7.V23.0.15.CDPredictRMS}{1.55}
\DefMacro{N7ToN7Final.CNN.transferFromN7V20P50.N7.V23.0.2.thresholdPredictRelativeRMS}{1.44}
\DefMacro{N7ToN7Final.CNN.transferFromN7V20P50.N7.V23.0.2.CDPredictRMS}{1.55}
\DefMacro{N7ToN7Final.CNN.transferFromN7V20P50.N7.V23.0.3.thresholdPredictRelativeRMS}{1.38}
\DefMacro{N7ToN7Final.CNN.transferFromN7V20P50.N7.V23.0.3.CDPredictRMS}{1.47}
\DefMacro{N7ToN7Final.CNN.transferFromN7V20P50.N7.V23.0.4.thresholdPredictRelativeRMS}{1.33}
\DefMacro{N7ToN7Final.CNN.transferFromN7V20P50.N7.V23.0.4.CDPredictRMS}{1.42}
\DefMacro{N7ToN7Final.CNN.transferFromN7V20P50.N7.V23.0.5.thresholdPredictRelativeRMS}{1.28}
\DefMacro{N7ToN7Final.CNN.transferFromN7V20P50.N7.V23.0.5.CDPredictRMS}{1.37}

\DefMacro{N7ToN7Final.ResNN.transferFromN7V20P50.N7.V23.0.01.thresholdPredictRelativeRMS}{1.52}
\DefMacro{N7ToN7Final.ResNN.transferFromN7V20P50.N7.V23.0.01.CDPredictRMS}{1.60}
\DefMacro{N7ToN7Final.ResNN.transferFromN7V20P50.N7.V23.0.05.thresholdPredictRelativeRMS}{1.34}
\DefMacro{N7ToN7Final.ResNN.transferFromN7V20P50.N7.V23.0.05.CDPredictRMS}{1.43}
\DefMacro{N7ToN7Final.ResNN.transferFromN7V20P50.N7.V23.0.1.thresholdPredictRelativeRMS}{1.30}
\DefMacro{N7ToN7Final.ResNN.transferFromN7V20P50.N7.V23.0.1.CDPredictRMS}{1.38}
\DefMacro{N7ToN7Final.ResNN.transferFromN7V20P50.N7.V23.0.15.thresholdPredictRelativeRMS}{1.27}
\DefMacro{N7ToN7Final.ResNN.transferFromN7V20P50.N7.V23.0.15.CDPredictRMS}{1.35}
\DefMacro{N7ToN7Final.ResNN.transferFromN7V20P50.N7.V23.0.2.thresholdPredictRelativeRMS}{1.23}
\DefMacro{N7ToN7Final.ResNN.transferFromN7V20P50.N7.V23.0.2.CDPredictRMS}{1.31}
\DefMacro{N7ToN7Final.ResNN.transferFromN7V20P50.N7.V23.0.3.thresholdPredictRelativeRMS}{1.16}
\DefMacro{N7ToN7Final.ResNN.transferFromN7V20P50.N7.V23.0.3.CDPredictRMS}{1.23}
\DefMacro{N7ToN7Final.ResNN.transferFromN7V20P50.N7.V23.0.4.thresholdPredictRelativeRMS}{1.09}
\DefMacro{N7ToN7Final.ResNN.transferFromN7V20P50.N7.V23.0.4.CDPredictRMS}{1.16}
\DefMacro{N7ToN7Final.ResNN.transferFromN7V20P50.N7.V23.0.5.thresholdPredictRelativeRMS}{1.01}
\DefMacro{N7ToN7Final.ResNN.transferFromN7V20P50.N7.V23.0.5.CDPredictRMS}{1.08}

\DefMacro{N7ToN7Final.ResNN.transferFromN7V20P5.N7.V23.0.01.thresholdPredictRelativeRMS}{1.94}
\DefMacro{N7ToN7Final.ResNN.transferFromN7V20P5.N7.V23.0.01.CDPredictRMS}{2.03}
\DefMacro{N7ToN7Final.ResNN.transferFromN7V20P5.N7.V23.0.05.thresholdPredictRelativeRMS}{1.67}
\DefMacro{N7ToN7Final.ResNN.transferFromN7V20P5.N7.V23.0.05.CDPredictRMS}{1.78}
\DefMacro{N7ToN7Final.ResNN.transferFromN7V20P5.N7.V23.0.1.thresholdPredictRelativeRMS}{1.50}
\DefMacro{N7ToN7Final.ResNN.transferFromN7V20P5.N7.V23.0.1.CDPredictRMS}{1.60}
\DefMacro{N7ToN7Final.ResNN.transferFromN7V20P5.N7.V23.0.15.thresholdPredictRelativeRMS}{1.41}
\DefMacro{N7ToN7Final.ResNN.transferFromN7V20P5.N7.V23.0.15.CDPredictRMS}{1.50}
\DefMacro{N7ToN7Final.ResNN.transferFromN7V20P5.N7.V23.0.2.thresholdPredictRelativeRMS}{1.32}
\DefMacro{N7ToN7Final.ResNN.transferFromN7V20P5.N7.V23.0.2.CDPredictRMS}{1.41}
\DefMacro{N7ToN7Final.ResNN.transferFromN7V20P5.N7.V23.0.3.thresholdPredictRelativeRMS}{-1.00}
\DefMacro{N7ToN7Final.ResNN.transferFromN7V20P5.N7.V23.0.3.CDPredictRMS}{-1.00}
\DefMacro{N7ToN7Final.ResNN.transferFromN7V20P5.N7.V23.0.4.thresholdPredictRelativeRMS}{-1.00}
\DefMacro{N7ToN7Final.ResNN.transferFromN7V20P5.N7.V23.0.4.CDPredictRMS}{-1.00}
\DefMacro{N7ToN7Final.ResNN.transferFromN7V20P5.N7.V23.0.5.thresholdPredictRelativeRMS}{-1.00}
\DefMacro{N7ToN7Final.ResNN.transferFromN7V20P5.N7.V23.0.5.CDPredictRMS}{-1.00}

\DefMacro{N7ToN7Final.ResNN.transferFromN7V20P10.N7.V23.0.01.thresholdPredictRelativeRMS}{1.82}
\DefMacro{N7ToN7Final.ResNN.transferFromN7V20P10.N7.V23.0.01.CDPredictRMS}{1.91}
\DefMacro{N7ToN7Final.ResNN.transferFromN7V20P10.N7.V23.0.05.thresholdPredictRelativeRMS}{1.57}
\DefMacro{N7ToN7Final.ResNN.transferFromN7V20P10.N7.V23.0.05.CDPredictRMS}{1.67}
\DefMacro{N7ToN7Final.ResNN.transferFromN7V20P10.N7.V23.0.1.thresholdPredictRelativeRMS}{1.51}
\DefMacro{N7ToN7Final.ResNN.transferFromN7V20P10.N7.V23.0.1.CDPredictRMS}{1.61}
\DefMacro{N7ToN7Final.ResNN.transferFromN7V20P10.N7.V23.0.15.thresholdPredictRelativeRMS}{1.43}
\DefMacro{N7ToN7Final.ResNN.transferFromN7V20P10.N7.V23.0.15.CDPredictRMS}{1.52}
\DefMacro{N7ToN7Final.ResNN.transferFromN7V20P10.N7.V23.0.2.thresholdPredictRelativeRMS}{1.34}
\DefMacro{N7ToN7Final.ResNN.transferFromN7V20P10.N7.V23.0.2.CDPredictRMS}{1.43}
\DefMacro{N7ToN7Final.ResNN.transferFromN7V20P10.N7.V23.0.3.thresholdPredictRelativeRMS}{-1.00}
\DefMacro{N7ToN7Final.ResNN.transferFromN7V20P10.N7.V23.0.3.CDPredictRMS}{-1.00}
\DefMacro{N7ToN7Final.ResNN.transferFromN7V20P10.N7.V23.0.4.thresholdPredictRelativeRMS}{-1.00}
\DefMacro{N7ToN7Final.ResNN.transferFromN7V20P10.N7.V23.0.4.CDPredictRMS}{-1.00}
\DefMacro{N7ToN7Final.ResNN.transferFromN7V20P10.N7.V23.0.5.thresholdPredictRelativeRMS}{-1.00}
\DefMacro{N7ToN7Final.ResNN.transferFromN7V20P10.N7.V23.0.5.CDPredictRMS}{-1.00}

\DefMacro{N7ToN7Final.CNN.N7.V23.ratio.thresholdPredictRelativeRMS}{1.00}
\DefMacro{N7ToN7Final.CNN.N7.V23.ratio.CDPredictRMS}{1.00}
\DefMacro{N10ToN7Final.CNN.transferFromN10P50.N7.V23.ratio.thresholdPredictRelativeRMS}{0.77}
\DefMacro{N10ToN7Final.CNN.transferFromN10P50.N7.V23.ratio.CDPredictRMS}{0.77}
\DefMacro{N10ToN7Final.ResNN.transferFromN10P50.N7.V23.ratio.thresholdPredictRelativeRMS}{0.70}
\DefMacro{N10ToN7Final.ResNN.transferFromN10P50.N7.V23.ratio.CDPredictRMS}{0.69}
\DefMacro{ActiveLearning.KMeans.ResNN.transferFromN10P50.N7.V23.ratio.thresholdPredictRelativeRMS}{0.63}
\DefMacro{ActiveLearning.KMeans.ResNN.transferFromN10P50.N7.V23.ratio.CDPredictRMS}{0.64}
\DefMacro{N7ToN7Final.CNN.transferFromN7V20P50.N7.V23.ratio.thresholdPredictRelativeRMS}{0.69}
\DefMacro{N7ToN7Final.CNN.transferFromN7V20P50.N7.V23.ratio.CDPredictRMS}{0.69}
\DefMacro{N7ToN7Final.ResNN.transferFromN7V20P50.N7.V23.ratio.thresholdPredictRelativeRMS}{0.60}
\DefMacro{N7ToN7Final.ResNN.transferFromN7V20P50.N7.V23.ratio.CDPredictRMS}{0.60}
\DefMacro{N7ToN7Final.ResNN.transferFromN7V20P5.N7.V23.ratio.thresholdPredictRelativeRMS}{0.69}
\DefMacro{N7ToN7Final.ResNN.transferFromN7V20P5.N7.V23.ratio.CDPredictRMS}{0.69}
\DefMacro{N7ToN7Final.ResNN.transferFromN7V20P10.N7.V23.ratio.thresholdPredictRelativeRMS}{0.69}
\DefMacro{N7ToN7Final.ResNN.transferFromN7V20P10.N7.V23.ratio.CDPredictRMS}{0.68}

\DefMacro{CNN.targetCD.1.5}{40}
\DefMacro{CNNFromN10P50.targetCD.1.5}{30}
\DefMacro{ResNetFromN10P50.targetCD.1.5}{15}
\DefMacro{KMeansResNetFromN10P50.targetCD.1.5}{10}
\DefMacro{CNNFromN7P50.targetCD.1.5}{30}
\DefMacro{ResNetFromN7P50.targetCD.1.5}{5}
\DefMacro{CNN.targetCD.1.75}{20}
\DefMacro{CNNFromN10P50.targetCD.1.75}{15}
\DefMacro{ResNetFromN10P50.targetCD.1.75}{5}
\DefMacro{KMeansResNetFromN10P50.targetCD.1.75}{5}
\DefMacro{CNNFromN7P50.targetCD.1.75}{5}
\DefMacro{ResNetFromN7P50.targetCD.1.75}{1}
\DefMacro{CNN.targetCD.2}{15}
\DefMacro{CNNFromN10P50.targetCD.2}{5}
\DefMacro{ResNetFromN10P50.targetCD.2}{5}
\DefMacro{KMeansResNetFromN10P50.targetCD.2}{5}
\DefMacro{CNNFromN7P50.targetCD.2}{1}
\DefMacro{ResNetFromN7P50.targetCD.2}{1}
\DefMacro{CNN.targetCD.2.25}{10}
\DefMacro{CNNFromN10P50.targetCD.2.25}{5}
\DefMacro{ResNetFromN10P50.targetCD.2.25}{5}
\DefMacro{KMeansResNetFromN10P50.targetCD.2.25}{1}
\DefMacro{CNNFromN7P50.targetCD.2.25}{1}
\DefMacro{ResNetFromN7P50.targetCD.2.25}{1}
\DefMacro{CNN.targetCD.2.5}{10}
\DefMacro{CNNFromN10P50.targetCD.2.5}{1}
\DefMacro{ResNetFromN10P50.targetCD.2.5}{1}
\DefMacro{KMeansResNetFromN10P50.targetCD.2.5}{1}
\DefMacro{CNNFromN7P50.targetCD.2.5}{1}
\DefMacro{ResNetFromN7P50.targetCD.2.5}{1}

\DefMacro{MinDataReduction}{2}
\DefMacro{MaxDataReduction}{10}

\DefMacro{TCADMinDataReduction}{3}
\DefMacro{TCADMaxDataReduction}{10}

%% file: abstract.tex
\begin{abstract}

Lithography simulation is one of the key steps in physical verification, enabled by the substantial optical and resist models.  
A resist model bridges the aerial image simulation to printed patterns. 
While the effectiveness of learning-based solutions for resist modeling has been demonstrated, they are considerably data-demanding.
Meanwhile, a set of manufactured data for a specific lithography configuration is only valid for the training of one single model, indicating low data efficiency. 
Due to the complexity of the manufacturing process, obtaining enough data for acceptable accuracy becomes very expensive in terms of both time and cost, 
especially during the evolution of technology generations when the design space is intensively explored. 
In this work, we propose a new resist modeling framework for contact layers, 
utilizing existing data from old technology nodes and active selection of data in a target technology node,
to reduce the amount of data required from the target lithography configuration. 
Our framework based on transfer learning and active learning techniques is effective within a competitive range of accuracy, 
i.e., \UseMacro{TCADMinDataReduction}-\UseMacro{TCADMaxDataReduction}X reduction on the amount of training data with comparable accuracy to the state-of-the-art learning approach. 

\end{abstract}

\iftrue
\begin{IEEEkeywords}
Lithography modeling, 
Machine learning, 
Transfer learning,
Active learning, 
Convolutional neural networks,
Residual neural networks
\end{IEEEkeywords}
\fi

%% file: intro.tex
\section{Introduction}
\label{sec:Introduction}

Due to the continuous semiconductor scaling from 10$nm$ technology node (N10) to 7$nm$ node (N7) \cite{DFM_SPIE2015_Liebmann, GDR_VLSIT2016_Liebmann}, 
the prediction of printed pattern sizes is becoming increasingly difficult and complicated due to the complexity of manufacturing process and variations. 
However, complex designs demand accurate simulations to guarantee functionality and yield.
Resist modeling, as a key component in lithography simulation, is critical to bridge the aerial image simulation to manufactured wafer data.
Rigorous simulations that perform physics-level modeling suffer from large computational overhead, which are not suitable when used extensively. 
Thus compact resist models are widely used in practice. 

\begin{figure}[tb]
\centering
\subfloat[]{\includegraphics[width=0.33\textwidth]{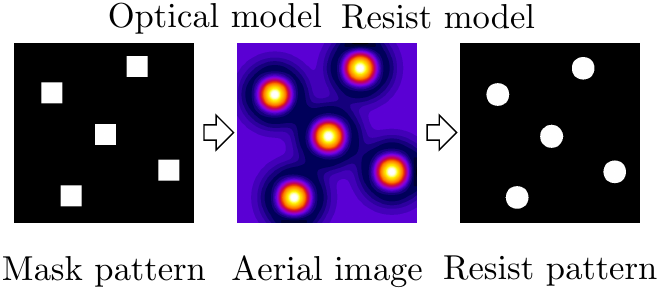}\label{fig:resistmodel}}
\subfloat[]{\includegraphics[width=0.14\textwidth]{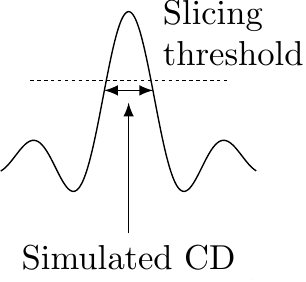}\label{fig:threshold}}
\caption{~\protect\subref{fig:resistmodel} Process of lithography simulation with optical and resist models. 
    \protect\subref{fig:threshold} Thresholds for aerial image determine simulated CD, which should match manufactured CD. }
\label{fig:LithoSimAndResistModel}
\end{figure}

Figure~\subref*{fig:resistmodel} shows the process of lithography simulations 
where the aerial image is computed from the input mask patterns and the optical model,  
and the output pattern is computed from the aerial image and the resist model. 
As the aerial image contains the light intensity map, the resist model needs to determine the slicing thresholds for the output patterns as shown in Figure~\subref*{fig:threshold}. 
With the thresholds, the critical dimensions (CDs) of printed patterns can be computed, which need to match CDs measured from manufactured patterns. 
In practice, various factors may impact a resist model such as the physical properties of photoresist, design rules of patterns, process variations. 
\textcolor{black}{
Critical dimension usually refers to the smallest dimension on a lithography level that must be accurately controlled when fabricating a device. 
Here critical dimensions refer to the sizes of printed patterns. 
}


Accurate lithography simulation like rigorous physics-based simulation is notorious for its long computational time, 
while simulation with compact models suffers from accuracy issues \cite{RESIST_SPIE2017_Shim, RESIST_SPIE2017_Watanabe}. 
On the other hand, machine learning techniques are able to construct accurate models and then make efficient predictions. 
These approaches first take training data to calibrate a model and then use this model to make predictions on testing data for validation. 
The effectiveness of learning-based solutions has been studied in various lithography related areas including 
aerial image simulation \cite{AERIAL_AO2017_Ma}, 
hotspot detection \cite{HOTSPOT_SPIE2011_Wuu, HOTSPOT_SPIE2016_Matsunawa, HOTSPOT_JM3-2016_Shin, HOTSPOT_DAC2017_Yang, HOTSPOT_SOCC2017_Yang, ML_PT2017_Lin}, 
optical proximity correction (OPC) \cite{OPC_TSM2008_Gu, OPC_JOP2010_Jia, OPC_JOIOP2013_Luo, OPC_JM32016_Matsunawa}, 
sub-resolution assist features (SRAF) \cite{MLSRAF_TCAD2017_Xu, SRAF_SPIE2015_ChinBoon}, 
resist modeling \cite{RESIST_SPIE2017_Shim, RESIST_SPIE2017_Watanabe}, etc.
In resist modeling, a convolutional neural network (CNN) that predicts slicing thresholds in aerial images is proposed \cite{RESIST_SPIE2017_Watanabe}. 
The neural network consists of three convolution layers and two fully connected layers. 
Since the slicing threshold is a continuous value, learning a resist model is a regression task rather than a classification task.  
Around 70\% improvement in accuracy is reported compared with calibrated compact models from Mentor Calibre \cite{TOOL_calibre}. 
Shim et al. \cite{RESIST_SPIE2017_Shim} propose an artificial neural network (ANN) with five hidden layers to predict the height of resist after exposure. 
Significant speedup is reported with high accuracy compared with a rigorous simulation. 

Although the learning-based approaches are able to achieve high accuracy, they are generally data-demanding in model training. 
In other words, big data is assumed to guarantee accuracy and generality. 
Furthermore, one data sample can only be used to train the corresponding model under the same lithography configuration, indicating a low data efficiency. 
Here data efficiency evaluates the accuracy a model can achieve given a specific amount of data,
or the amount of data samples are required to achieve target accuracy. 
Nevertheless, obtaining a large amount of data is often expensive and time-consuming, 
especially when the technology node switches from one to another and the design space is under active exploration, e.g., from N10 to N7.  
The lithography configurations including optical sources, resist materials, etc., are frequently changed for experiments.  
Therefore, a fast preparation of models with high accuracy is urgently desired. 
In addition, it remains to be a question that what are the best designs for building a model. 
Typical practice of regular array patterns or random patterns may not be representative enough to calibrate accurate and generic models. 
Thus effective techniques to recognize representative designs will also be beneficial to improving data efficiency.  

Different from previous approaches, in this work, we assume the availability of large amounts of data from the previous technology generation with old lithography configurations
and small amounts of data from a target lithography configuration.  
We focus on increasing the data efficiency by 
1) reusing those from other lithography configurations and transfer the knowledge between different configurations; 
2) active selection of data samples in the target configuration, also known as active learning. 
The objective is to achieve accurate resist models with significantly fewer data to a target configuration.  
The major contributions are summarized as follows. 
\begin{itemize}
    \item We propose a high performance resist modeling technique based on the residual neural network (ResNet). 
    \item We propose a transfer learning scheme for ResNet that can reduce the amount of data with a target accuracy by utilizing the data from other configurations. 
    \item We propose an active learning scheme based on K-Medoids algorithm with theoretical insights for both CNN and ResNet. 
    \item The experimental results demonstrate \UseMacro{TCADMinDataReduction}-\UseMacro{TCADMaxDataReduction}X reduction in the amount of training data to achieve accuracy comparable to the state-of-the-art learning approach \cite{RESIST_SPIE2017_Watanabe}. 
\end{itemize}

The rest of the paper is organized as follows. 
Section~\ref{sec:Preliminary} illustrates the problem formulation. 
Section~\ref{sec:Algorithms} explains the details of our approach. 
The effectiveness of our approach is verified in Section~\ref{sec:ExperimentalResults} and the conclusion is drawn in Section~\ref{sec:Conclusion}.

%% file: prelim.tex
\section{Preliminaries}
\label{sec:Preliminary}

In this section, we will briefly introduce the background knowledge on lithography simulation and resist modeling. 
Then the problem formulation is explained. 
We mainly focus on contact layers in this work, but our methodology shall be applicable to other layers.
For simplicity, we use the word \textit{label} to represent the target value for prediction, e.g,. threshold, given a data sample; 
we also use the phrase \textit{unlabelled} data to denote data samples whose labels are unknown. 

\subsection{Lithography Simulation}
\label{sec:LithographySimulation}

Lithography simulation is generally composed of two stages, i.e., optical simulation and resist simulation, where optical and resist models are required, respectively. 
In the optical simulation, an optical model, characterized by the illumination tool, takes mask patterns to compute aerial images, i.e., light intensity maps. 
Then in the resist simulation, a resist model finalizes the resist patterns with the aerial images from the optical simulation. 
Generally, there are two types of resist models. 
One is a variable threshold resist (VTR) model in which the thresholds vary according to aerial images, 
and the other is a constant threshold resist (CTR) model in which the light intensity is modulated in an aerial image.  
We adopt the former since it is suitable to learning-based approaches \cite{RESIST_SPIE2017_Watanabe}. 

\begin{figure}
    \centering
    \subfloat[]{\includegraphics[width=0.091\textwidth]{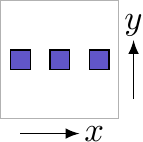}\label{fig:AerialImageContacts}}
    \hspace{.2in}
    \subfloat[]{\includegraphics[width=0.25\textwidth]{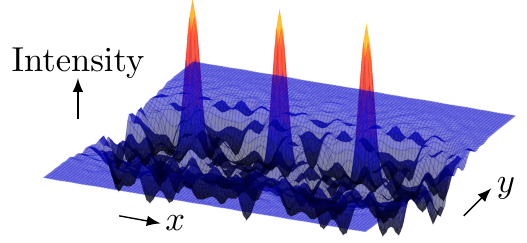}\label{fig:AerialImage}}\\
    \subfloat[]{\includegraphics[width=0.1\textwidth]{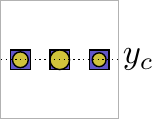}\label{fig:AerialImageContactsContour}}
    \hspace{.15in}
    \subfloat[]{\includegraphics[width=0.25\textwidth]{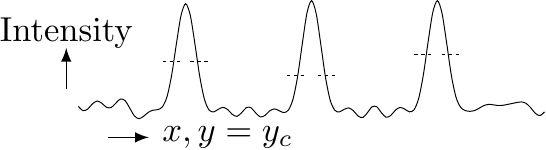}\label{fig:AerialImageProfile}}
    \caption{~\protect\subref{fig:AerialImageContacts} Design target of 3 contacts and \protect\subref{fig:AerialImage} the light intensity plot of aerial image. 
    Assume that RETs such as SRAF and OPC have been already applied to the contacts before optical simulation. 
    \protect\subref{fig:AerialImageContactsContour} A dotted line horizontally crosses the centers at $y=y_c$ and the circles denote the contours of printed patterns.  
    \protect\subref{fig:AerialImageProfile} Light intensity profiling along the dotted line at $y=y_c$ extracted from the aerial image and different slicing thresholds for each contact. 
}
\label{fig:AerialImageToThreshold}
\end{figure}

Figure~\ref{fig:AerialImageToThreshold} shows an example of lithography simulation for a clip with three contacts. 
We assume that proper resolution enhancement techniques (RETs) such as OPC and SRAF have been applied before the computation of the aerial image \cite{RET_IBM2001_Liebmann}. 
The optical simulation generates the aerial image, as shown in Figure~\subref*{fig:AerialImage}. 
Resist simulation then computes the thresholds in the aerial image to predict printed patterns. 
If we want to measure the widths of contacts along the dotted line in Figure~\subref*{fig:AerialImageContactsContour}, 
the light intensity profiling can be extracted from the aerial image along the line and calculates the CDs for each contact with the thresholds. 

\subsection{Historical Data and Transfer Learning}
\label{sec:HistoricalData}

Since the lithography configurations evolve from one generation to another with the advancement of technology nodes, 
there are plenty of historical data available for the old generation. 
As mentioned in Section~\ref{sec:Introduction}, accurate models require a large amount of data for training or calibration, 
which are expensive to obtain during the exploration of a new generation. 
If the lithography configurations have no fundamental changes, 
the knowledge learned from the historical data may still be applicable to the new configuration, 
which can eventually help to reduce the amount of new data required. 

Transfer learning represents a set of techniques to transfer the knowledge from one or multiple source domains to a target domain, 
utilizing the underlying similarity between the data from these domains. 
Various studies have explored the effectiveness of knowledge transfer in image recognition and robotics \cite{ML_AAAI2017_Hanna, ML_ARXIV2016_Rusu, ML_TKDE2010_Pan}, 
while it is not clear whether the knowledge between different resist models is transferable or not. 

\begin{figure}[tb]
    \centering 
    \subfloat[]{\includegraphics[width=0.1\textwidth]{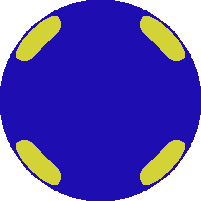}\label{fig:OpticalSourceN10}}
    \hspace{.2in}
    \subfloat[]{\includegraphics[width=0.1\textwidth]{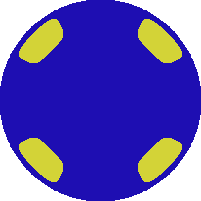}\label{fig:OpticalSourceN7}}
    \caption{~Optical sources (yellow) for \protect\subref{fig:OpticalSourceN10} N10 and \protect\subref{fig:OpticalSourceN7} N7.
    }
    \label{fig:OpticalSource}
\end{figure}

In this work, we consider the evolution of the contact layer from the cutting edge technology node N10 to N7 \cite{DFM_SPIE2015_Liebmann, GDR_VLSIT2016_Liebmann}. 
A large amount of available N10 data are assumed. 
During the evolution to N7, different design rules for mask patterns, optical sources and resist materials for lithography are explored. 
Table~\ref{tab:LithographyConfigurations} shows the lithography configurations considered for N10 and N7.   
Differences in letters $A, B$ represent different configurations of design rules, optical sources, or resist materials. 
One configuration for N10 is considered, 
while two configurations are considered for N7, i.e., $\textrm{N7}_a$, $\textrm{N7}_b$, with two kinds of resist materials (about 20\% difference in the slopes of dissolution curves). 
From N10 to N7, both the design rules and optical sources are changed. 
For N10, we consider a pitch of 64$nm$ with double patterning lithography, 
while for N7, the pitch is set to 45$nm$ with triple patterning lithography \cite{DFM_SPIE2015_Liebmann}. 
The width of each contact is set to half pitch. 
The lithography target of each contact is set to 60$nm$ for both N10 and N7. 
Optical sources calibrated with industrial strength for N10 and N7 are shown in Figure~\ref{fig:OpticalSource}, with the same type of illumination shapes. 

Various combinations of knowledge transfer can be explored from Table~\ref{tab:LithographyConfigurations}, 
such as N10$\rightarrow$N7, $\textrm{N7}_i$$\rightarrow$$\textrm{N7}_j$, and N10$+$$\textrm{N7}_i$$\rightarrow$$\textrm{N7}_j$, where $i \ne j, i, j \in \{a, b\}$. 

\begin{table}[tb]
\centering
\caption{Lithography Configurations for N10 and N7}
\label{tab:LithographyConfigurations}
\begin{tabular}{|c|c|cc|}
\hline
\multirow{2}{*}{} & \multicolumn{1}{c|}{\multirow{2}{*}{N10}} & \multicolumn{2}{c|}{N7}                                         \\ \cline{3-4} 
                  & \multicolumn{1}{c|}{}                     & \multicolumn{1}{c|}{$\textrm{N7}_a$} & $\textrm{N7}_b$ \\ \hline
Design Rule       & $A$                                         & $B$                          & $B$     \\ \hline
Optical Source    & $A$                                         & $B$                          & $B$     \\ \hline
Resist Material   & $A$                                         & $A$                          & $B$     \\ \hline
\end{tabular}
\end{table}

\subsection{Active Learning for Regression}
\label{sec:ActiveLearning}

Active learning assumes unlabelled data samples exist in a pool or can be generated. 
Querying for data labels is very expensive and the amount of queries should be minimized. 
Thus selecting the proper and limited portion of data samples for querying is essential to modeling accuracy. 

We define the problem of pool-based active learning as: 
given a pool of unlabelled data samples, select $k$ samples to query for labels and train a model to maximize the accuracy across the entire dataset. 
Be aware that the selection of data samples should not depend on data labels since labels are unknown before querying. 
Hence, active learning is very effective in improving data efficiency without the requirement of any additional labelled data. 

There are extensive studies for active learning in classification for CNN and SVM \cite{ML_ARXIV2017_Sener, ML_JMLR2001_Tong, ML_ICML2015_Berlind, ML_ARXIV2016_Donahue}. 
A few studies have explored active learning for support vector regression (SVR) and multi-layer perception (MLP) \cite{ML_PR2014_Demir, ML_TNN2000_Fukumizu}. 
Most techniques are categorized into confidence level or clustering approaches. 
Confidence level approaches tend to choose data samples with low prediction confidence, 
and clustering approaches choose representative subset of data samples among an entire dataset. 
There are also successful applications of active learning in VLSI CAD related areas \cite{DFM_ICCAD2010_Zhuo, ANALOG_ICCAD2012_Lin, ICCAM_TCAD2017_Li}. 

However, practical studies on active learning techniques for regression tasks with CNN or ResNet are lacking, 
and its performance when combined with transfer learning is unclear. 
It is often difficult to evaluate the confidence level with large and complicated models like CNN or ResNet, 
while clustering approaches only rely on the general properties of data samples and models. 
Therefore, we explore effective clustering strategies for active selection of data samples, which are suitable to regression tasks with CNN and ResNet.  

\subsection{Learning-based Resist Modeling}
\label{sec:ResistModeling}

\begin{figure}[tb!]
    \centering \subfloat[]{\includegraphics[width=0.08\textwidth]{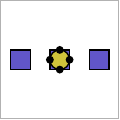}\label{fig:contact4Edges}}
    \hfill
    \subfloat[]{\includegraphics[width=0.08\textwidth]{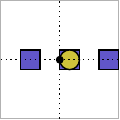}\label{fig:contactLeftEdge}}
    \hfill
    \subfloat[]{\includegraphics[width=0.08\textwidth]{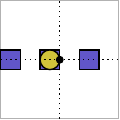}\label{fig:contactRightEdge}}
    \hfill
    \subfloat[]{\includegraphics[width=0.08\textwidth]{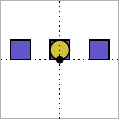}\label{fig:contactBelowEdge}}
    \hfill
    \subfloat[]{\includegraphics[width=0.08\textwidth]{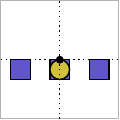}\label{fig:contactAboveEdge}}
    \hfill
    \caption{~\protect\subref{fig:contact4Edges} The thresholds for the middle of the 4 edges of the center contact are predicted.
        \protect\subref{fig:contactLeftEdge}
        \protect\subref{fig:contactRightEdge}
        \protect\subref{fig:contactBelowEdge}
        \protect\subref{fig:contactAboveEdge}
        The clip window is shifted such that the target position lies in the center of the clip. 
    }
    \label{fig:contactEdges}
\end{figure}

The thresholds of positions near the contacts are of significant importance since they usually determine the boundaries of printed contacts. 
Hence we consider the middle of the left, right, bottom and top edges for each contact, 
as shown in Figure~\subref*{fig:contact4Edges}, 
where the positions for prediction are highlighted with black dots. 
As the threshold is mainly influenced by the surrounding mask patterns,  
resist models typically compute the threshold using a clip of mask patterns centered by a target position. 
To measure the thresholds in Figure~\subref*{fig:contact4Edges}, 
we select a clip where the target position lies in its center, as shown in Figure~\subref*{fig:contactLeftEdge} to Figure~\subref*{fig:contactAboveEdge}. 
The task of a resist model is to compute the thresholds for these positions of each contact \cite{RESIST_SPIE2017_Watanabe}. 

Learning-based resist modeling consists of two phases, i.e., training and testing. 
In the training phase, training dataset with both aerial images and thresholds are used to calibrate the model, 
while in the testing phase, the model predicts thresholds for the aerial images from the testing dataset. 

\subsection{Problem Formulation}
\label{sec:ProblemFormulation}


The accuracy\footnote{\scriptsize{Note that the accuracy we talk about in this paper refers to the accuracy at end of lithography flow including all RETs.}} 
of a model is evaluated with root mean square (RMS) error defined as follows, 
\begin{equation}
    \epsilon = \sqrt{\frac{1}{N} \sum_{i=1}^{N} (\hat{y}-y)^2}, 
    \label{eq:RMSError}
\end{equation}
where $N$ denotes the amount of samples, $y$ denotes the golden values and $\hat{y}$ denotes the predicted values. 
We further define relative RMS error, 
\begin{equation}
    \epsilon_r = \sqrt{\frac{1}{N} \sum_{i=1}^{N} (\frac{\hat{y}-y}{y})^2}, 
    \label{eq:RelativeRMSError}
\end{equation}
where a relative ratio of error from the golden values can be represented. 
Both metrics can refer to errors in either CD or threshold. 
Although during model training, the RMS error of threshold is generally minimized due to easier computation, 
the eventual model is often evaluated with the RMS error of CD for its physical meaning to the patterns. 
The RMS errors in threshold and CD essentially have almost the same fidelity, and usually yield consistent comparison. 
For convenience, we report relative RMS error in threshold ($\epsilon_r^{th}$) for comparison of different models since it removes the dependency to the scale of thresholds, 
and use RMS error in CD ($\epsilon^{CD}$) for data efficiency related comparison. 

\begin{definition}[Data Efficiency]
    The amount of target domain data required to learn a model with a given accuracy. 
\end{definition}
Given a specific amount of data from a target domain, if one can learn a model with a higher accuracy than another, it also indicates higher data efficiency. 
Thus improving model accuracy benefits data efficiency as well. 

The resist modeling problem is defined as follows. 
\begin{problem}[Learning-based Resist Modeling]
    Given a dataset containing information of aerial images and thresholds at their centers, train a resist model that can maximize the accuracy for the prediction of thresholds. 
\end{problem}
In practice, accuracy is not the only objective. 
The amount of training data should be minimized as well due to the high cost of data preparation.  
Therefore, we propose the problem of data efficient resist modeling as follows. 
\begin{problem}[Data Efficient Resist Modeling]
    Given a labelled N10 dataset containing aerial images and thresholds, and an unlabelled N7 dataset containing aerial images only, 
    train a resist model for target dataset $\textrm{N7}_i$ that can achieve high accuracy 
    and meanwhile query labels for as few $\textrm{N7}_i$ data samples as possible, where $i \in \{a, b\}$.  
\end{problem}
Minimizing the times of label querying is equivalent to minimizing the cost of data preparation, 
since the most expensive part is to obtain the labels, i.e., thresholds, through either manufactured wafer data or rigorous simulation.

%% file: algo.tex
\section{Algorithms}
\label{sec:Algorithms}

\begin{figure}
    \centering
    \includegraphics[width=0.35\textwidth]{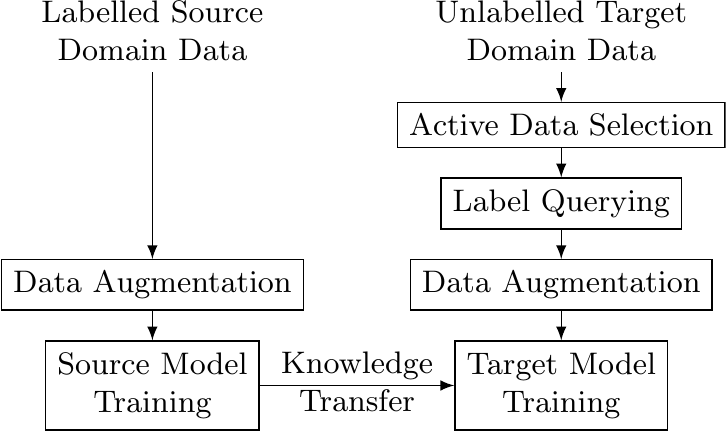}
    \caption{~Training flow with transfer learning and active learning.}
    \label{fig:TrainFlow}
\end{figure}

In this section, we will explain the structure of our models and then the details regarding the transfer learning and active learning schemes. 
Figure~\ref{fig:TrainFlow} shows the overall training flow. 
We first leverage labelled source domain data to train a source domain model. 
Then before training the target domain model, active learning is applied for active selection of data samples for label querying. 
The target domain model is eventually trained with selected data samples and knowledge transferred from the source domain model. 
Data augmentation in Section~\ref{sec:DataAugmentation} is applied before training of both source and target models. 

\subsection{Data Preparation}
\label{sec:DataPreparation}

\begin{figure}
    \includegraphics[width=0.45\textwidth]{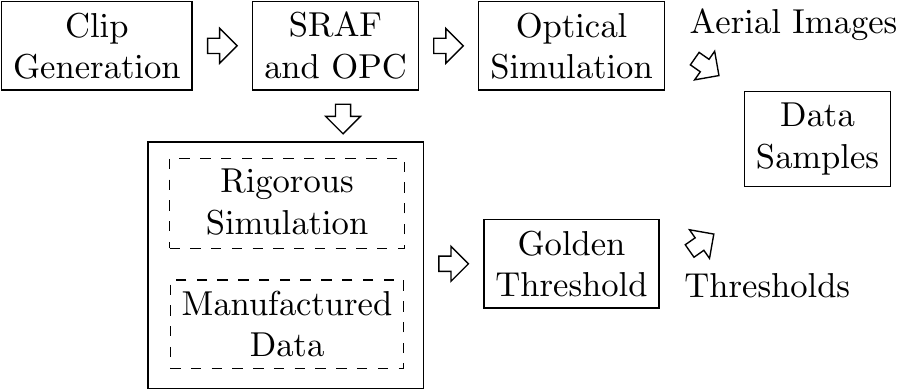}
    \caption{~Flow of data preparation.}
    \label{fig:DataPreparationFlow}
\end{figure}

Figure~\ref{fig:DataPreparationFlow} gives the flow of data preparation. 
We first generate clips and perform SRAF insertion and OPC. 
The aerial images are then computed from the optical simulation, 
and at the same time, the golden thresholds need to be computed from either the rigorous simulation or the manufactured data. 
Each data sample consists of an aerial image and the threshold at its center. 

\subsubsection{Clip Generation}
\label{sec:RandomClipGeneration}

Following the design rules such as minimum pitch of contacts, we generate three types of $2 \times 2 \mu m$ clips. 
It is necessary to ensure that there is a contact in the center of each clip since that is the target contact for threshold computation. 

\begin{figure}
\centering 
\subfloat[]{\includegraphics[]{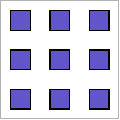}\label{fig:ContactArray}}
\hspace{.1in}
\subfloat[]{\includegraphics[]{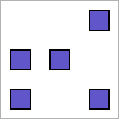}\label{fig:RandomizedContactArray}}
\hspace{.1in}
\subfloat[]{\includegraphics[]{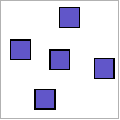}\label{fig:RandomizedContactPosition}}
\caption{~\protect\subref{fig:ContactArray} A clip of $3 \times 3$ contact array. 
    \protect\subref{fig:RandomizedContactArray} A clip of $3 \times 3$ randomized contact array. 
    \protect\subref{fig:RandomizedContactPosition} A clip of contacts with random positions. 
}
\end{figure}

\textbf{Contact Array}. 
All possible $m \times n$ arrays of contacts within the dimensions of clips are enumerated.  
The steps of the arrays can be multiple times of the minimum pitch $p$, i.e., $p, 2p, 3p, \dots, $ in horizontal or vertical directions. 
An example of $3 \times 3$ contact array with a certain pitch is shown in Figure~\subref*{fig:ContactArray}. 
It needs to mention that the same $3 \times 3$ contact array with different steps should be regarded as different clips due to discrepant spacing. 

\textbf{Randomized Contact Array}. 
The aforementioned contact arrays essentially distribute contacts on grids and fill all the slots in the grid maps. 
The randomization of contact arrays is implemented by a random distribution of contacts in those grid maps. 
Figure~\subref*{fig:RandomizedContactArray} shows an example of randomized contact array from the $3 \times 3$ contact array in Figure~\subref*{fig:ContactArray}. 
Various distribution of contacts can be generated even from the same grid maps. 

\textbf{Contacts with Random Positions}. 
Contacts in this type of clips do not necessarily align to any grid map, as their positions are randomly generated,
while the design rules are still guaranteed. 
An example is shown in Figure~\subref*{fig:RandomizedContactPosition}. 
No matter how the surrounding contacts change, the contact in the center of the clip should remain the same. 

\subsubsection{Data Augmentation}
\label{sec:DataAugmentation}

\begin{figure}
\centering 
\subfloat[]{\includegraphics[width=0.062\textwidth]{DataAugment}\label{fig:DataAugment1}}
\subfloat[]{\includegraphics[width=0.062\textwidth]{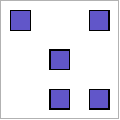}\label{fig:DataAugment2}}
\subfloat[]{\includegraphics[width=0.062\textwidth]{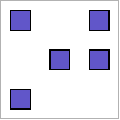}\label{fig:DataAugment3}}
\subfloat[]{\includegraphics[width=0.062\textwidth]{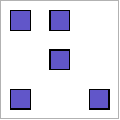}\label{fig:DataAugment4}}
\subfloat[]{\includegraphics[width=0.062\textwidth]{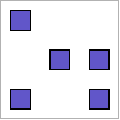}\label{fig:DataAugment5}}
\subfloat[]{\includegraphics[width=0.062\textwidth]{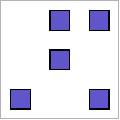}\label{fig:DataAugment6}}
\subfloat[]{\includegraphics[width=0.062\textwidth]{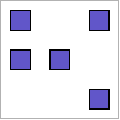}\label{fig:DataAugment7}}
\subfloat[]{\includegraphics[width=0.062\textwidth]{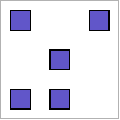}\label{fig:DataAugment8}}
\caption{~Combinations of rotation and flipping. 
    \protect\subref{fig:DataAugment1} Original. 
    \protect\subref{fig:DataAugment2} Rotate $90^{\circ}$.
    \protect\subref{fig:DataAugment3} Rotate $180^{\circ}$. 
    \protect\subref{fig:DataAugment4} Rotate $270^{\circ}$. 
    \protect\subref{fig:DataAugment1} Flip. 
    \protect\subref{fig:DataAugment2} Flip and rotate $90^{\circ}$. 
    \protect\subref{fig:DataAugment3} Flip and rotate $180^{\circ}$. 
    \protect\subref{fig:DataAugment4} Flip and rotate $270^{\circ}$. 
}
\label{fig:DataAugment}
\end{figure}

Due to the symmetry of optical sources in Figure~\ref{fig:OpticalSource}, data can be augmented with rotation and flipping, improving the data efficiency \cite{ML_BOOK2016_Goodfellow}. 
Eight combinations of rotation and flipping are shown in Figure~\ref{fig:DataAugment}, where new data samples are obtained without new thresholds. 
Data augmentation inflates datasets to obtain models with better generalization. 

\subsection{Convolutional Neural Networks}
\label{sec:CNN}

Convolutional neural networks (CNN) have demonstrated impressive performance on mask related applications in lithography 
such as hotspot detection, and resist modeling \cite{HOTSPOT_DAC2017_Yang, RESIST_SPIE2017_Watanabe}. 
The structure of CNN mainly includes convolution layers and fully connected layers.  
Features are extracted from convolution layers and then classification or regression is performed by fully connected layers. 
Figure~\subref*{fig:CNN} illustrates a CNN structure with three convolution layers and two fully connected layers \cite{RESIST_SPIE2017_Watanabe}.  
The first convolution layer has 64 filters with dimensions of $7 \times 7$. 
Although not explicitly shown most of the time, a rectified linear unit (ReLU) layer for activation is applied immediately after the convolution layer, 
where the ReLU function is defined as, 
\textcolor{black}{
\begin{equation}
    f(x^{l-1}) = \left\{
        \begin{array}{ll}
            x^{l-1}, \quad & \textrm{if } x^{l-1} \ge 0, \\
            0, \quad       & \textrm{otherwise}. 
        \end{array}
        \right.
        \label{eq:relu}
\end{equation}
}
Then the max-pooling layer performs down-sampling with a factor of 2 to reduce the feature dimensions and improve the invariance to translation \cite{ML_BOOK2016_Goodfellow}. 
After three convolution layers, two fully connected layers are applied 
where the first one has 256 hidden units followed with a ReLU layer and a 50\% dropout layer, and second one connects to the output. 

\subsection{Residual Neural Networks}
\label{sec:ResNet}

\begin{figure}[tb]
\centering
\vspace{-.1in}
\begin{tikzpicture}[every node/.style={inner sep=0pt, outer sep=0pt}]
\begin{axis}[
hide axis, 
xmin=0, xmax=10, 
ymin=0.01, ymax=0.02, 
legend style={font=\small, at={(1.05, 0.5)}, anchor=west, draw=none}, 
legend columns=2, 
]
\addlegendimage{mark=none, orange}
\addlegendentry{5-layer}
\addlegendimage{mark=none, myblue}
\addlegendentry{10-layer}
\end{axis}
\end{tikzpicture}\vspace{-.0in}\\
\subfloat[]{\includegraphics[width=0.22\textwidth]{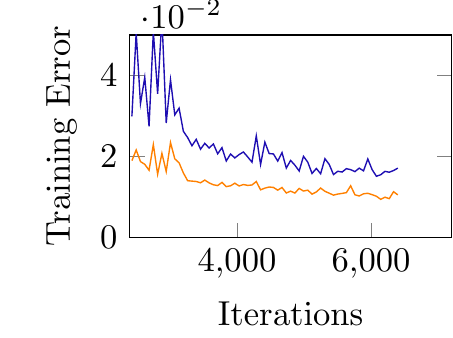}\label{fig:TrainCurves}}
\hspace{.1in}
\subfloat[]{\includegraphics[width=0.22\textwidth]{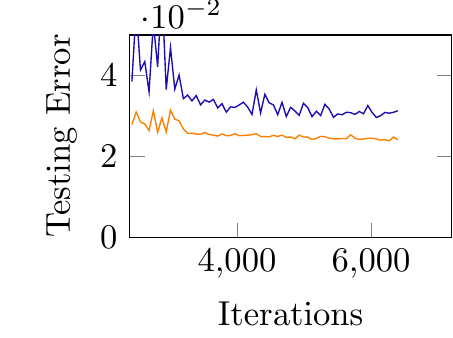}\label{fig:TestCurves}}
\caption{~Counterintuitive \protect\subref{fig:TrainCurves} training and \protect\subref{fig:TestCurves} testing errors for different depth of CNN with epochs.}
\label{fig:TrainTestCurves}
\end{figure}

One way to improve the performance of CNN is to increase the depth for a larger capacity of the neural networks. 
However, the counterintuitive degradation of training accuracy in CNN is observed when stacking more layers, 
preventing the neural networks from better performance \cite{ML_CVPR2016_He}. 
An example of CNNs with 5 and 10 layers is shown in Figure~\ref{fig:TrainTestCurves}, 
where the deeper CNN fails to converge to a smaller training error than the shallow one due to gradient vanishing \cite{ML_TNN1994_Bengio, ML_ICAIS2010_Glorot}, 
eventually resulting in the failure to achieve a better testing error either. 
The study from He et al. \cite{ML_CVPR2016_He} reveals that the underlying reason comes from the difficulty of identity mapping. 
In other words, fitting a hypothesis $\mathcal{H}(x) = x$ is considerably difficult for solvers to find optimal solutions. 
To overcome this issue, residual neural networks (ResNet), which utilizes shortcut connections, are adopted to assist the convergence of training accuracy. 

\begin{figure}[tb!]
\centering
\includegraphics[width=0.2\textwidth]{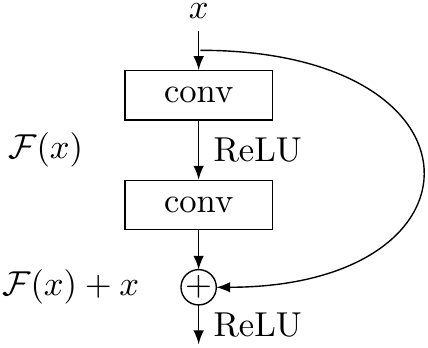}
\caption{Building block of ResNet.}
\label{fig:ResNNBlock}
\vspace{-.1in}
\end{figure}

The building block of ResNet is illustrated in Figure~\ref{fig:ResNNBlock}, 
where a shortcut connection is inserted between the input and output of two convolution layers. 
Let the function $\mathcal{F}(x)$ be the mapping defined by the two convolution layers. 
Then the entire function for the building block becomes $\mathcal{F}(x)+x$. 
Suppose the building block targets to fit the hypothesis $\mathcal{H}(x)$. 
The residual networks train $\mathcal{F}(x) = \mathcal{H}(x) - x$, 
while the convolution layers without shortcut connections like that in CNN try to directly fit $\mathcal{F}(x) = \mathcal{H}(x)$. 
Theoretically, if $\mathcal{H}(x)$ can be approximated with $\mathcal{F}(x)$, 
then it can also be approximated with $\mathcal{F}(x)+x$. 
Despite the same nature, comprehensive experiments have demonstrated a better convergence of ResNet than that of CNN for deep neural networks \cite{ML_CVPR2016_He}. 
We also observe a better performance of ResNet with the transfer learning schemes than that of CNN in our problem, which has never been explored before. 

\begin{figure}[tb!]
\centering
\subfloat[]{\includegraphics[]{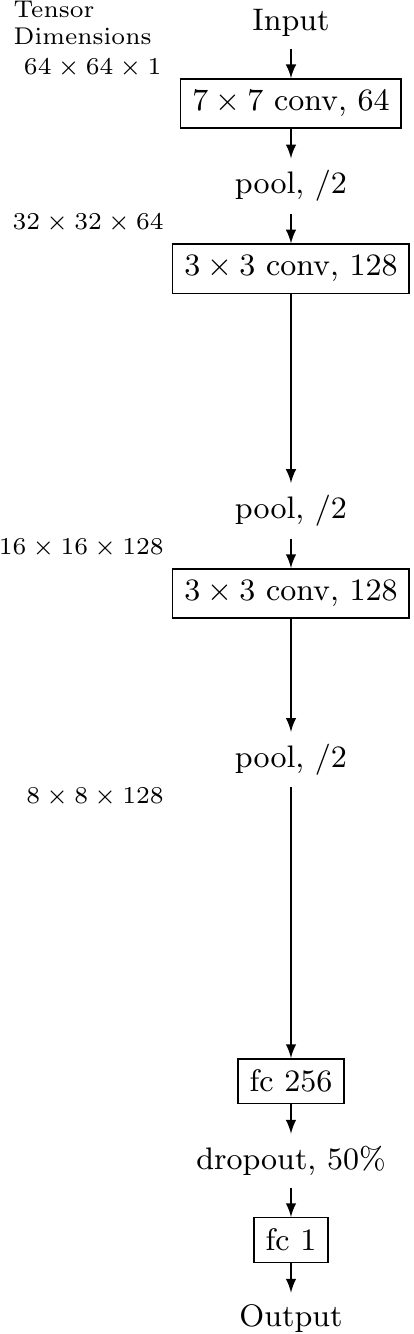}\label{fig:CNN}}
\hspace{.2in}
\subfloat[]{\includegraphics[]{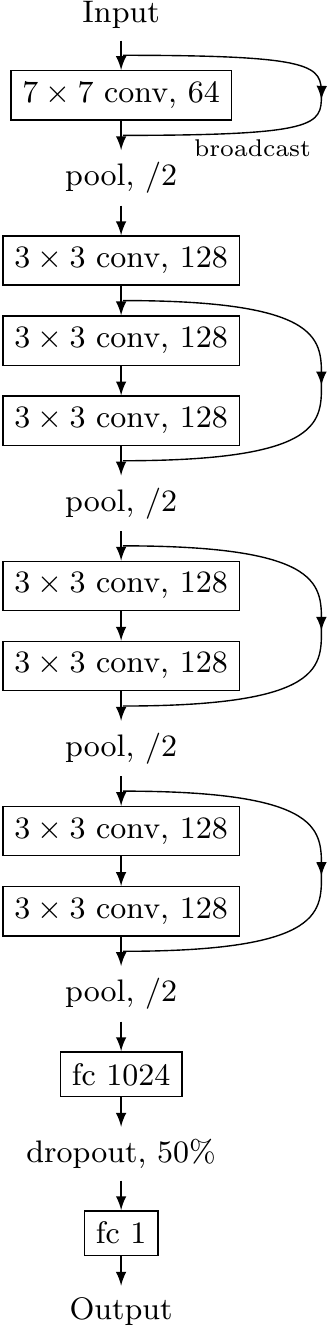}\label{fig:ResNN}}
\caption{~\protect\subref{fig:CNN} CNN and \protect\subref{fig:ResNN} ResNet structure.}
\label{fig:NeuralNetworks}
\end{figure}

The ResNet is shown in Figure~\subref*{fig:ResNN} with 8 convolution layers and 2 fully connected layers.  
Different from the original setting \cite{ML_CVPR2016_He}, we add a shortcut connection to the first convolution layer 
by broadcasting the input tensor of $64 \times 64 \times 1$ to $64 \times 64 \times 64$. 
This minor change enables better empirical results in our problem. 
For the rest of the networks, 3 building blocks for ResNet are utilized. 

\subsection{Transfer Learning}
\label{sec:TransferLearning}

Transfer learning aims at adapting the knowledge learned from data in source domains to a target domain. 
The transferred knowledge will benefit the learning in the target domain with a faster convergence and better generalization \cite{ML_BOOK2016_Goodfellow}. 
Suppose the data in the source domain has a distribution $P_s$ and that in the target domain has a distribution $P_t$. 
The underlying assumption of transfer learning lies in the common factors that need to be captured for learning the variations of $P_s$ and $P_t$, 
so that the knowledge for $P_s$ is also useful for $P_t$. 
An intuitive example is that learning to recognize cats and dogs in the source task helps the recognition of ants and wasps in the target task, 
especially when the source task has significantly larger dataset than that of the target task. 
The reason comes from the low-level notions of edges, shapes, etc., shared by many visual categories \cite{ML_BOOK2016_Goodfellow}. 
In resist modeling, different lithography configurations can be viewed as separate tasks with different distributions. 

\begin{figure}
    \centering
    \includegraphics[]{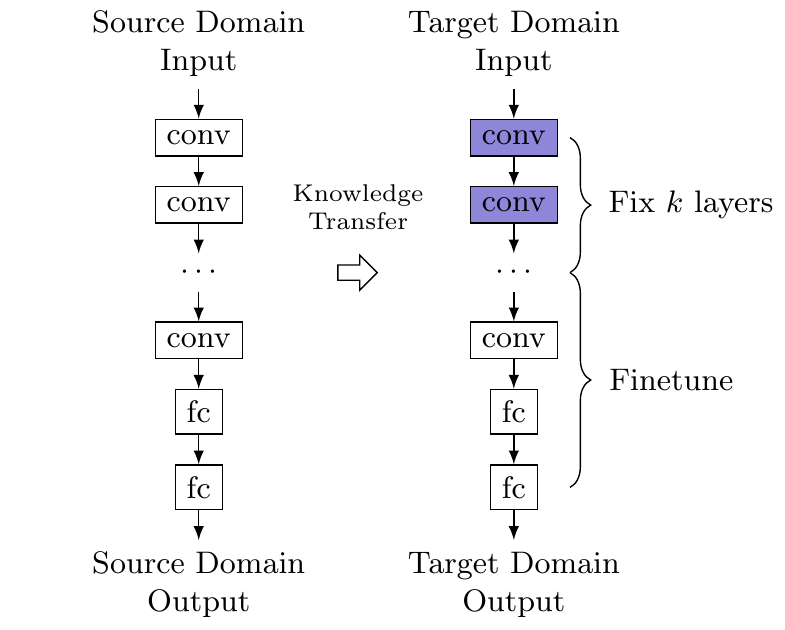}
    \caption{~Transfer learning scheme with the first $k$ layers fixed when training for target domain, denoted as $\textrm{TF}_k$.}
    \label{fig:Transfer}
\end{figure}

Typical transfer learning scheme for neural networks fixes the first several layers of the model trained for another domain 
and finetune the successive layers with data from the target domain. 
The first several layers usually extract general features, which are considered to be similar between the source and the target domains, 
while the successive layers are classifiers or regressors that need to be adjusted. 
Figure~\ref{fig:Transfer} shows an example of the transfer learning scheme.
We first train a model with source domain data and then use the source domain model as the starting point for the training of the target domain. 
During the training for the target domain, the first $k$ layers are fixed, while the rest layers are finetuned. 
We denote this scheme as $\textit{TF}_k$, shortened from ``Transfer and Fix'', where $k$ is the parameter for the number of fixed layers. 

\textcolor{black}{
In this work, we focus on the impacts of transfer learning and do not consider various preprocessing steps like scaling and normalization. 
In other words, raw aerial images are fed to the neural networks. 
The benefits of scaling and normalization are left to future work. 
}

\subsection{Active Learning with Clustering}
\label{sec:ActiveLearningWithCoreSet}

Although transfer learning is potentially able to improve the accuracy of the target dataset using knowledge from a source dataset, 
selection of representative target data samples may further improve the accuracy. 
Let $D$ be the unlabeled dataset in the target domain and $\bm{s}$ be the set of selected data samples for label querying, where $|\bm{s}| \le k$ and $k$ is the maximum number of data samples for querying. 
For any $(\bm{x}_i, y_i) \in D$, $\bm{x}_i$ is the feature, e.g., aerial image, and $y_i$ is the label, e.g., threshold, where $y_i$ is unknown for $D$. 
Consider a loss function $l(\bm{x}_i, y_i; \bm{w})$ parameterized over the hypothesis class ($\bm{w}$), e.g., parameters of a learning algorithm. 
The objective of active learning is to minimize the average loss of dataset $D$ with a model trained from $\bm{s}$, 
\begin{equation}
    \min_{\bm{s}: |\bm{s}| \le k, \bm{s} \in D } \frac{1}{n} \sum_{i = 1}^{n} l(\bm{x}_i, y_i; \bm{w}_{\bm{s}}) , 
    \label{eq:ActiveLearningObj}
\end{equation}
where $n = |D|$, and $\bm{w}_{\bm{s}}$ represents the parameters of a model trained from $\bm{s}$. 

We present an upper bound of Eq.~\eqref{eq:ActiveLearningObj} for any Lipschitz loss function and Lipschitz estimator. 
Then we show that both CNN and ResNet with non-linear ReLU activations are actually Lipschitz continuous. 
We also assume the training loss can drop to zero, which is likely to be achieved with large enough models. 

{\color{black}
\begin{definition}\label{def:g_lipschitz}
Let $g(\cdot;\cdot) : \mathbb{R}^{d} \times \mathbb{R}^{d} \rightarrow \mathbb{R}$, we say $g$ is $L_1$-Lipschitz continuous with respect to $g(*;\cdot)$ if
\begin{align}
| g({\bm{x}};\bm{w}) - g({\bm{x}}';\bm{w}) | \leq L_1 \cdot \| {\bm{x}} - {\bm{x}}' \|. \notag 
\end{align}
\end{definition}
We also write $g({\bm{x}};\bm{w})$ as $g_{\bm{w}}({\bm{x}})$.
We use Frobenius norm for norm of a matrix here, i.e., $\norm{\cdot}$. 

\begin{definition}\label{def:f_lipschitz}
Let $f(\cdot, \cdot; \cdot) : \mathbb{R}^{d_1} \times \mathbb{R} \times \mathbb{R}^{d_2} \rightarrow \mathbb{R}_{\geq 0}$, we say $f$ is $L_2$-Lipschitz continuous with respect to $f(*, * ;\cdot)$ if
\begin{align}
| f({\bm{x}},y;\bm{w}) - f({\bm{x}}',y';\bm{w}) | \leq L_2 \cdot ( \| {\bm{x}} - {\bm{x}}'\| + |  y - y' |), \notag \\
\forall {\bm{x}} \in \mathbb{R}^{d_1}, \forall y, y' \in \mathbb{R}, \forall \bm{w} \in \mathbb{R}^{d_2}. \notag
\end{align}
\end{definition}
We also write $f({\bm{x}},y;\bm{w})$ as $f_{\bm{w}}({\bm{x}},y)$.
}

We state the following theorem: 

\begin{theorem}
    Given $n$ \textcolor{black}{independent and identically distributed (i.i.d.) random samples} as $D = \{\bm{x}_i, y_i\}_{i \in \{1, 2, \ldots, n\}}$, and a set of selected points $\bm{s}$. 
    If the following properties hold, 
    \begin{enumerate}
        \item loss function $l(\bm{x}, y; \bm{w})$ is $\lambda^l$-Lipschitz continuous w.r.t $(\bm{x}, y)$; 
        \item the ground truth of label $y = f(\bm{x}) + \epsilon$ has the property that $f(\cdot)$ is $\lambda^f$-Lipschitz continuous and random noise $\epsilon \sim \mathcal{N}(0, \sigma^2)$; 
        \item $\hat{f}(\cdot)$ in the prediction function $\hat{y} = \hat{f}(\bm{x})$ is $\lambda^{\hat{f}}$-Lipschitz; 
        \item $l(\bm{x}_j, y_j; \bm{w}_{\bm{s}}) = 0, \forall j \in \bm{s}$, where $\bm{w}_{\bm{s}}$ is the weights of the trained model with samples $\bm{s}$; 
    \end{enumerate}
    then we have the following inequality, 
    \begin{equation}
        \frac{1}{n}\sum_{i \in D} l(\bm{x}_i, y_i; \bm{w}_{\bm{s}}) 
        \le \frac{\lambda^l(\lambda^f + 1)}{n} \sum_{j \in \bm{s}} \sum_{i=1}^{k_j} \norm{ \bm{x}_i - \bm{x}_j^c } 
        + 2\lambda^l \sum_{i \in D} \abs{ \epsilon_i },  
        \label{eq:AvgLossBound}
    \end{equation}
    where \textcolor{black}{
        $k_j$ is the number of samples whose closest sample in $\bm{s}$ is $\bm{x}_j^c$;  
    }
    $\abs{\epsilon_i}$ is a sample from an independent random half-normal distribution with mean $\frac{\sigma \sqrt{2}}{\sqrt{\pi}}$ and variance $\sigma^2(1-\frac{2}{\pi})$.  
    \label{theorem:AvgLossBound}
\end{theorem}
The left-hand side of the inequality is the average loss across the entire dataset. 
The right-hand side, i.e., the upper bound of the average loss, is correlated to the objective of a \textit{K-Medoids Clustering} problem \cite{CLUSTER_EML2010_Jin}, 
\textcolor{black}{
where K is the number of labeled data samples for training ($K = |\bm{s}|$). 
K-Medoids clustering problem is required to return K clusters from a set of points as well as K centers for each cluster. 
}
Therefore, minimizing $\sum_{j \in \bm{s}} \sum_{i=1}^{k_j} \norm{\bm{x}_i - \bm{x}_j}$ helps to bound the left-hand side. 

\begin{figure}
    \centering 
    \subfloat[]{\includegraphics[width=0.22\textwidth]{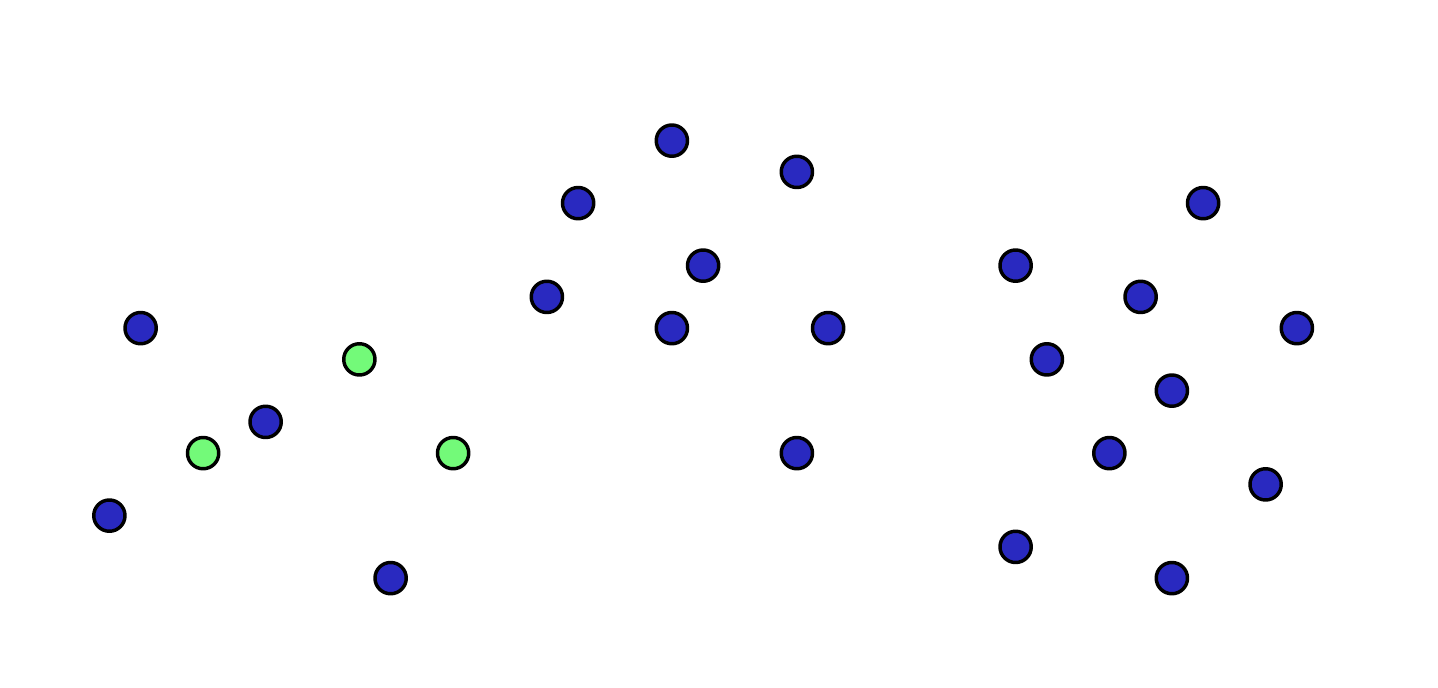}}{\label{fig:random}}
    \subfloat[]{\includegraphics[width=0.22\textwidth]{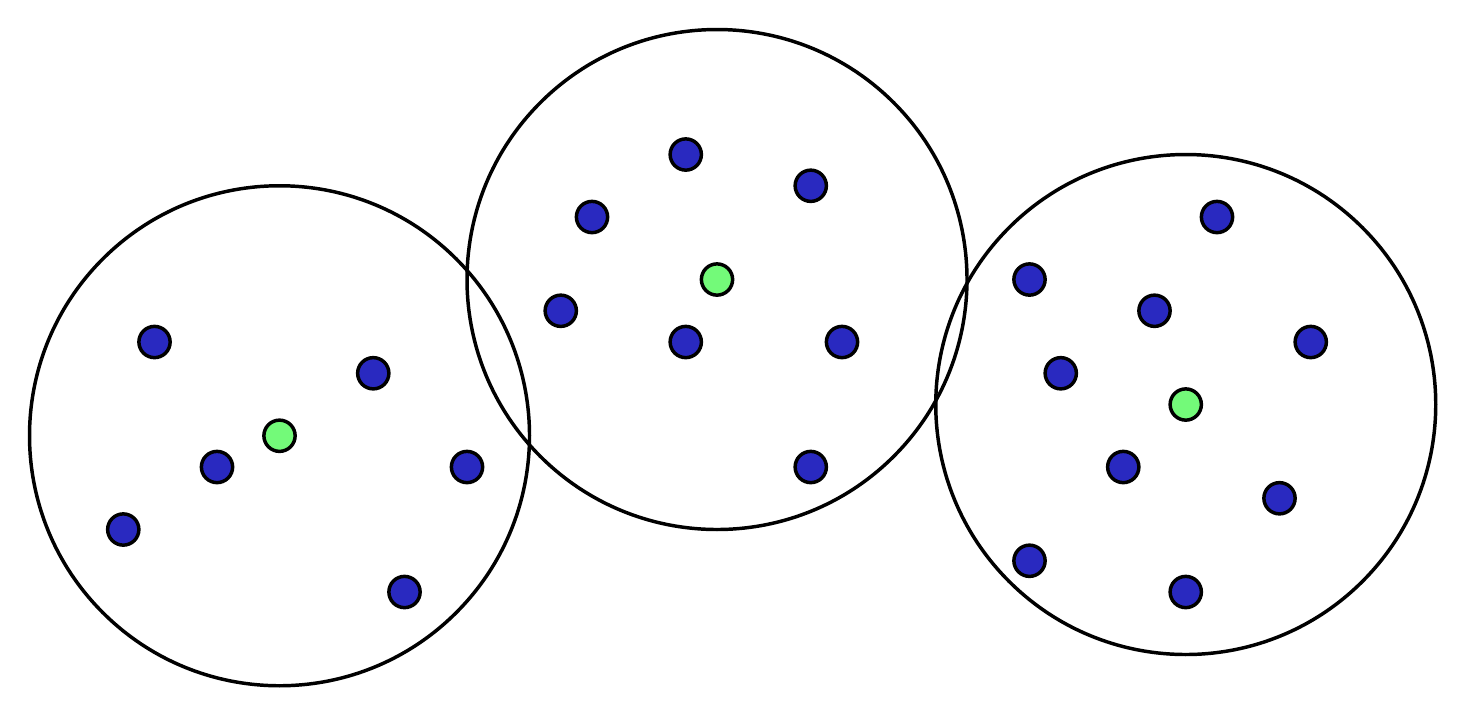}}{\label{fig:kmedoids}}
    \caption{~Example of (a) bad data selection and (b) K-Medoids clustering selection in 2D space. Three selected points are highlighted. 
        Circles denote three clusters centered by selected points. 
    }
    \label{fig:ActiveLearningCluster}
\end{figure}
Figure~\ref{fig:ActiveLearningCluster} provides an intuition for the K-Medoids clustering in a 2D space. 
Random selection may result in biased coverage of the entire dataset, causing significant overfitting of model training. 
K-Medoids clustering is able to select medoids (data) evenly from the space for training such that most unselected data samples are close to their nearest medoids. 

Theorem~\ref{theorem:AvgLossBound} requires both the loss function and the estimator to be Lipschitz continuous. 
{\color{black}
\begin{lemma}
    If the following conditions hold, 
    \begin{enumerate}
        \item $\forall i \in D, (\bm{x}_i, y_i) $ satisfies $ \norm{\bm{x}_i} \le b_1, \abs{y_i} \le b_2$; 
        \item $\hat{f}_{\bm{w}_{\bm{s}}}(\bm{x})$ is $\lambda^{\hat{f}}$-Lipschitz continuous w.r.t $\bm{x}$; 
        \item $\exists (\bm{x}_0, y_0)$ such that $\hat{f}_{\bm{w}_{\bm{s}}}(\bm{x}_0) = y_0 + \delta$, where $\delta$ is a bounded constant;
    \end{enumerate}
    then square loss function $l_{\bm{w}_{\bm{s}}}(\bm{x}, y) = (y - \hat{f}_{\bm{w}_{\bm{s}}}(\bm{x}))^2$ is $\lambda^{l}$-Lipschitz continuous, 
    where $y$ is the label and $\hat{f}_{\bm{w}_{\bm{s}}}(*)$ is the learned function with parameter $\bm{w}_{\bm{s}}$ (also denoted as $\hat{f}(*)$ for brevity), 
    \begin{equation}
        \lambda^{l} = ( 4 \lambda^{\hat{f}} b_1 + 4 b_2  + 2 \abs{\delta} ) \cdot \max(1, \lambda^{\hat{f}}). 
    \end{equation}
    \label{lemma:LossLipschitz}
\end{lemma}
In practice, the three assumptions are not difficult to hold. 
Consider the physical meaning of $\bm{x}$ and $y$, both $\norm{\bm{x}}$ and $\abs{y}$ are numerically small in this work. 
Lemma~\ref{lemma:CNNLipschitz} proves that CNN/ResNet is Lipschitz continuous. 
If the training error for CNN/ResNet is small, which is mostly true, $(\bm{x}_0, y_0)$ can be selected from the training dataset and then $\abs{\delta}$ is also small. 
}
\begin{lemma}
    A CNN/ResNet for regression with $n_c$ convolution layers (with max-pooling and ReLU) and $n_{fc}$ fully connected layers is $(1+\alpha\sqrt{N})^{n_c+n_{fc}}$-Lipschitz. 
    \label{lemma:CNNLipschitz}
\end{lemma}
Detailed proofs for Theorem~\ref{theorem:AvgLossBound}, Lemma~\ref{lemma:LossLipschitz}, and Lemma~\ref{lemma:CNNLipschitz} can be found in Appendix. 


%% file: result.tex
\section{Experimental Results}
\label{sec:ExperimentalResults}

Our framework is implemented with Tensorflow \cite{TOOL_TENSORFLOW} and validated on a Linux server with 3.4GHz Intel i7 CPU and Nvidia GTX 1080 GPU. 
The K-Medoids clustering is approximated by K-Means clustering in scikit-learn \cite{TOOL_sklearn} and assigning the data points that are closest to centroids as medoids. 
We observe that this approach provides better and more stable objectives of K-Medoids clustering than does dedicated K-Medoids clustering solver in PyClust package in our experiments. 

Around 980 mask clips are generated according to Section~\ref{sec:DataPreparation} for N10 and N7 separately following the design rules in Section~\ref{sec:HistoricalData}, respectively.  
$\textrm{N7}_a$ and $\textrm{N7}_b$ use the same set of clips, but different lithography configurations. 
SRAF, OPC and aerial image simulation are performed with Mentor Calibre \cite{TOOL_calibre}. 
The golden CD values are obtained from rigorous simulation using Synopsys Sentaurus Lithography models \cite{TOOL_slitho} calibrated from manufactured data 
for N10, $\textrm{N7}_a$, and $\textrm{N7}_b$ according to Table~\ref{tab:LithographyConfigurations}. 
Then golden thresholds are extracted. 
Each clip has four thresholds as shown in Figure~\ref{fig:contactEdges}. 
Hence the N10 dataset contains 3928 samples and each N7 dataset contains 3916 samples, respectively. 
The data augmentation technique in Section~\ref{sec:DataAugmentation} is applied, so the training set and the testing set will be augmented by a factor of 8 independently. 
For example, if 50\% of the data for N10 are used for training, then there are $3928 \times 50\% \times 8 = 15712$ samples. 
It needs to mention that always the same 50\% portions are used during the validation of a dataset for fair comparison of different techniques. 
The batch size is set to 32 for training accommodating to the large variability in the sizes of training datasets. 
Adam \cite{ML_ARXIV2014_Kingma} is used as the stochastic optimizer and maximum epoch is set to 200 for training. 

The training time for one model takes 10 to 40 minutes according to the portions of a dataset used for training, 
and prediction time for an entire N10 or N7 dataset takes less than 10 seconds, while the rigorous simulation takes more than 15 hours for each N10 or N7 dataset. 
Thus we no longer report the prediction time which is negligible compared with that of the rigorous simulation. 
Each experiment runs 10 different random seeds and averages the numbers. 

\subsection{CNN and ResNet}
\label{sec:CNNandResNet}

\begin{figure}[tb!]
\centering
\hspace{.2in}
\vspace{-.02in}
\begin{tikzpicture}[every node/.style={inner sep=0pt, outer sep=0pt}]
\begin{axis}[
hide axis, 
xmin=0, xmax=10, 
ymin=0.01, ymax=0.02, 
legend style={font=\small, at={(1.05, 0.5)}, anchor=west, draw=none}, 
legend columns=3, 
]
\addlegendimage{mark=*, black}
\addlegendentry{CNN-5}
\addlegendimage{mark=triangle*, orange, fill opacity=0.7}
\addlegendentry{CNN-10}
\addlegendimage{mark=triangle*, blue, opacity=0.7}
\addlegendentry{ResNet}
\addlegendimage{mark=pentagon*, flora, fill opacity=0.7}
\addlegendentry{CNN-5+AL\footnotemark}
\addlegendimage{mark=pentagon*, sky, opacity=0.7}
\addlegendentry{ResNet+AL\footnotemark[\value{footnote}]}
\end{axis}
\end{tikzpicture}\vspace{-.0in}\\
\subfloat[]
{
    \includegraphics[]{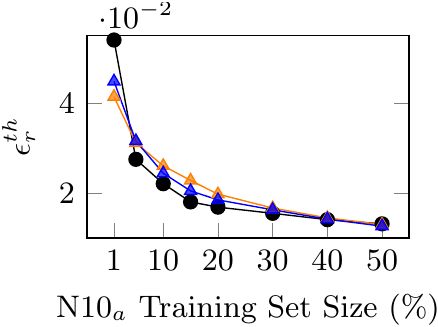}\label{fig:NTenCNNAlexNNResNN}
}
\subfloat[]
{
    \includegraphics[]{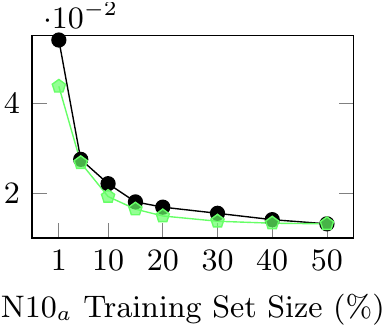}\label{fig:N10CNNAL}
}
\\
\subfloat[]
{
    \includegraphics[]{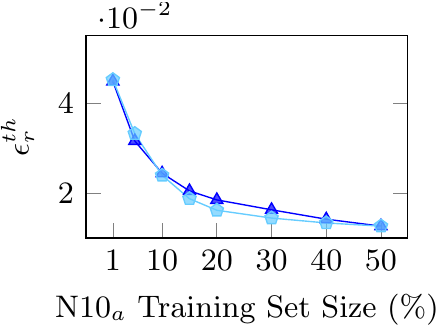}\label{fig:N10ResNNAL}
}
\subfloat[]
{
    \includegraphics[]{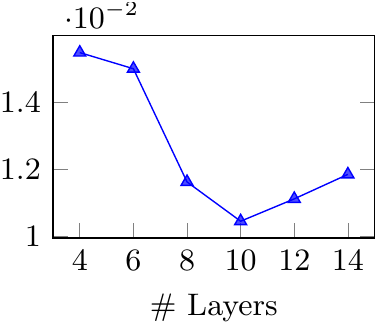}\label{fig:NSevenResNNBlocks}
}
\caption{
\protect\subref{fig:NTenCNNAlexNNResNN} Comparison on testing accuracy of CNN-5, CNN-10, and ResNet on N10. 
\protect\subref{fig:N10CNNAL} Testing accuracy of CNN with active learning on N10. 
\protect\subref{fig:N10ResNNAL} Testing accuracy of ResNet with active learning on N10. 
\protect\subref{fig:NSevenResNNBlocks} Impact of depth on the testing accuracy of ResNet. 
}
\label{fig:NNTestP50}
\end{figure}

We first compare CNN and ResNet in Figure~\subref*{fig:NTenCNNAlexNNResNN}.  
Column ``CNN-5'' denotes the network with 5 layers shown in Figure~\subref*{fig:CNN}. 
Column ``CNN-10'' denotes the one with 10 layers that has the same structure as that in Figure~\subref*{fig:ResNN} but without shortcut connections.  
Column ``ResNet'' denotes the one with 10 layers shown in Figure~\subref*{fig:ResNN}.
When using 1\% to 20\% training data, ResNet shows better average relative RMS error $\epsilon_r^{th}$ than CNN-10, but CNN-5 provides the best error. 
We will show later that ResNet on the contrary outperforms CNN-5 when transfer learning is incorporated. 

We then show the performance of active learning for CNN and ResNet in Figure~\subref*{fig:N10CNNAL} and Figure~\subref*{fig:N10ResNNAL}, denoted as ``CNN-5+AL'' and ``ResNet+AL'', respectively. 
The beneficial amount of training data for active data selection is from 10\% to 40\%. 
For example, 
for 20\% training data, it provides 11.6\% accuracy improvement for CNN and 12.5\% for ResNet; 
for 30\% training data, it provides 11.4\% improvement for CNN and 11.3\% for ResNet \cite{DFM_ISPD2018_Lin}. 
The benefit of active learning is not significant for extremely small training dataset, e.g., 1\% and 5\%.  
When there are very few training data, 
it is more likely for randomly selected data samples to distribute quite some distance away than to squeeze as small clusters. 
Although active selection of data can avoid corner cases of extremely poor sampling, e.g., all data samples squeezing as a small cluster, 
while it is difficult to demonstrate the benefit of active learning in ordinary cases. 
On the other hand, when the amount of training data increases, the benefit from active learning drops due to sufficient coverage. 
The rightmost points take all 50\% training data and thus show the same accuracy as that without active learning. 


The impacts of depth on the performance of ResNet are further explored in Figure~\subref*{fig:NSevenResNNBlocks}, 
where we gradually stack more building blocks in Figure~\ref{fig:ResNNBlock} before fully connected layers. 
The x-axis denotes total number of convolution and fully connected layers corresponding to different numbers of building blocks. 
For instance, 0 building block leads to 4 layers and 3 building blocks result in 10 layers (Figure~\subref*{fig:ResNN}). 
The testing error decreases to lowest value at 10 layers and then starts to increase, indicating potential overfitting afterwards \cite{ML_BOOK2016_Goodfellow}. 
Therefore, we use 10 layers for the ResNet in the experiment. 

\footnotetext{Results for active learning extended from \cite{DFM_ISPD2018_Lin}}


\subsection{Knowledge Transfer From N10 to N7}
\label{sec:TransferFromN10ToN7}

\begin{figure}[tb!]
\centering
\hspace{.2in}
\vspace{-.05in}
\begin{tikzpicture}[every node/.style={inner sep=0pt, outer sep=0pt}]
\begin{axis}[
hide axis, 
xmin=0, xmax=10, 
ymin=0.01, ymax=0.02, 
legend style={font=\small, at={(1.05, 0.5)}, anchor=west, draw=none}, 
legend columns=3, 
]
\addlegendimage{mark=*, black}
\addlegendentry{CNN}
\addlegendimage{mark=triangle*, orange, fill opacity=0.7}
\addlegendentry{CNN $\textrm{TF}_0$}
\addlegendimage{mark=triangle*, blue, opacity=0.7}
\addlegendentry{ResNet $\textrm{TF}_0$}
\addlegendimage{mark=triangle*, mygreen}
\addlegendentry{ResNet $\textrm{TF}_4$}
\addlegendimage{mark=triangle*, aspar}
\addlegendentry{ResNet $\textrm{TF}_6$}
\addlegendimage{mark=triangle*, flora}
\addlegendentry{ResNet $\textrm{TF}_8$}
\addlegendimage{mark=pentagon*, sky, opacity=0.7}
\addlegendentry{ResNet $\textrm{TF}_0$+AL\footnotemark[\value{footnote}]}
\end{axis}
\end{tikzpicture}\vspace{-.0in}\\
\subfloat[]
{
    \includegraphics[]{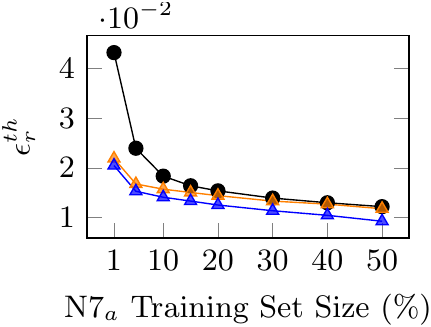}\label{fig:NTenToNSevenPNN}
}
\subfloat[]
{
    \includegraphics[]{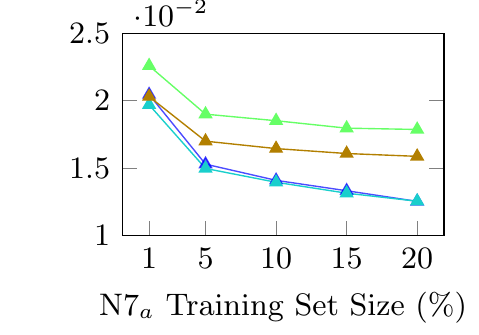}\label{fig:NSevenResNNTransferX}
}
\\
\subfloat[]
{
    \includegraphics[]{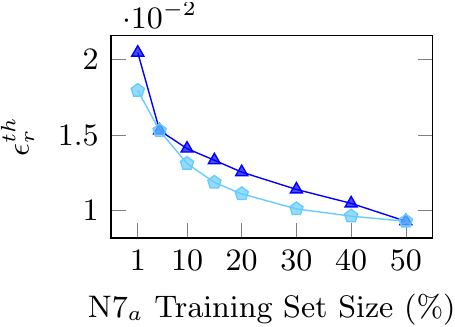}\label{fig:NTenToNSevenAL}
}
\caption{
Testing accuracy of transfer learning from N10 to $\textrm{N7}_a$.
\protect\subref{fig:NTenToNSevenPNN} Comparison between CNN and transfer learning. 
\protect\subref{fig:NSevenResNNTransferX} Comparison between transfer learning schemes where different numbers of layers are fixed. 
\protect\subref{fig:NTenToNSevenAL} Comparison between transfer learning only and transfer learning plus active learning for ResNet. 
}
\label{fig:NTenToNSeven}
\end{figure}

We then compare the testing accuracy between knowledge transfer from N10 to N7 and directly training from N7 datasets in Figure~\subref*{fig:NTenToNSevenPNN}. 
In this example, the x-axis represents the percentage of training dataset for the target domain $\textrm{N7}_a$, 
while the percentage of data from the source domain N10 is always 50\%. 
Similar trends are also observed for $\textrm{N7}_b$. 
Curve ``CNN'' denotes training the CNN of 5 layers in Figure~\subref*{fig:CNN} with data from target domain only, i.e., no transfer learning involved. 
Curve ``CNN $\textrm{TF}_0$'' denotes the transfer learning scheme in Section~\ref{sec:TransferLearning} for the same CNN with zero layer fixed. 
Curve ``ResNet $\textrm{TF}_0$'' denotes applying the same scheme to ResNet. 
The most significant benefit of transfer learning comes from small training dataset with a range of 1\% to 20\%, 
where there are around \UseMacro{N10ToN7Final.ResNN.transferFromN10P50.CNN.0.01.imp}\% to \UseMacro{N10ToN7Final.ResNN.transferFromN10P50.CNN.0.2.imp}\% improvement in the accuracy from CNN. 
Meanwhile, ResNet $\textrm{TF}_0$ can achieve an average of \UseMacro{N10ToN7Final.ResNN.transferFromN10P50.CNN.transferFromN10P50.avg}\% smaller error than CNN $\textrm{TF}_0$.  

Figure~\subref*{fig:NSevenResNNTransferX} further compares the results of fixing different numbers of layers during transfer learning. 
In this case, ResNet $\textrm{TF}_0$ and ResNet $\textrm{TF}_4$ have the best accuracy, while the error increases with more layers fixed. 
It is indicated that the tasks N10 and N7 are quite different and both feature extraction layers and regression layers need finetuning. 

In Figure~\subref*{fig:NTenToNSevenAL}, we enable transfer learning plus active learning, 
which provides 7\% to 11\% additional accuracy improvement for 10\% to 40\% amount of training data from the target domain. 

\subsection{Knowledge Transfer within N7}
\label{sec:TransferwithinN7}

\begin{figure}[tb!]
\centering
\hspace{.2in}
\vspace{-.05in}
\begin{tikzpicture}[every node/.style={inner sep=0pt, outer sep=0pt}]
\begin{axis}[
hide axis, 
xmin=0, xmax=10, 
ymin=0.01, ymax=0.02, 
legend style={font=\small, at={(1.05, 0.5)}, anchor=west, draw=none}, 
legend columns=3, 
]
\addlegendimage{mark=*, black}
\addlegendentry{CNN}
\addlegendimage{mark=diamond*, orange, fill opacity=0.7}
\addlegendentry{CNN $\textrm{TF}_0$}
\addlegendimage{mark=diamond*, blue, opacity=0.7}
\addlegendentry{ResNet $\textrm{TF}_0$}
\addlegendimage{mark=diamond*, mygreen}
\addlegendentry{ResNet $\textrm{TF}_4$}
\addlegendimage{mark=diamond*, aspar}
\addlegendentry{ResNet $\textrm{TF}_6$}
\addlegendimage{mark=diamond*, flora}
\addlegendentry{ResNet $\textrm{TF}_8$}
\end{axis}

\end{tikzpicture}\vspace{-.0in}\\
\subfloat[]
{
    \includegraphics[]{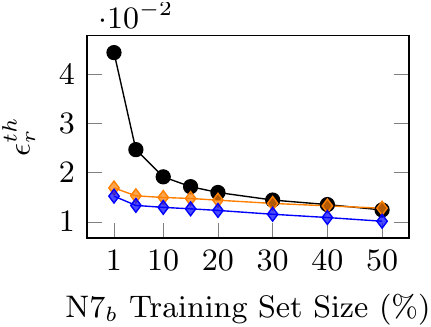}\label{fig:NSevenTrP50}
}
\subfloat[]
{
    \includegraphics[]{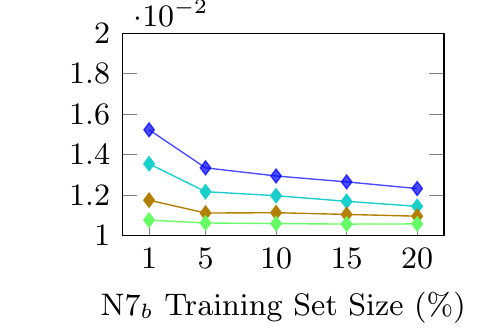}\label{fig:TransferFixXFromN7P50}
}
\caption{~Testing accuracy of transfer learning from $\textrm{N7}_a$ to $\textrm{N7}_b$.
\protect\subref{fig:NSevenTrP50} Comparison between CNN and transfer learning. 
\protect\subref{fig:TransferFixXFromN7P50} Comparison between transfer learning schemes where different numbers of layers are fixed. 
}
\label{fig:NSevenFromN7P50}
\end{figure}

The transfer learning between different N7 datasets, e.g., from $\textrm{N7}_a$ to $\textrm{N7}_b$, is also explored in Figure~\ref{fig:NSevenFromN7P50}. 
The x-axis represents the percentage of training dataset for the target domain $\textrm{N7}_b$, 
while the percentage of data from the source domain $\textrm{N7}_a$ is always 50\%. 
Compared with the knowledge transfer from N10 to N7, we achieve even higher accuracy between 1\% and 20\% training datasets in Figure~\subref*{fig:NSevenTrP50}. 
For example, with 1\% training dataset, there is around \UseMacro{N7ToN7Final.ResNN.transferFromN7V20P50.CNN.0.01.imp}\% improvement in accuracy from CNN, 
and with 20\% training dataset, the improvement is around \UseMacro{N7ToN7Final.ResNN.transferFromN7V20P50.CNN.0.2.imp}\%. 
ResNet $\textrm{TF}_0$ keeps having lower errors than that of CNN $\textrm{TF}_0$ as well, with an average benefit around \UseMacro{N7ToN7Final.ResNN.transferFromN7V20P50.CNN.transferFromN7V20P50.avg}\%. 


The curves in Figure~\subref*{fig:TransferFixXFromN7P50} show different insights from that of the knowledge transfer from N10 to N7. 
The accuracy of ResNet $\textrm{TF}_0$ can be further improved with more layers fixed, e.g., ResNet $\textrm{TF}_8$, by around \UseMacro{N7ToN7Final.ResNN.transferFix8FromN7V20P50.ResNN.transferFromN7V20P50.max}\% to \UseMacro{N7ToN7Final.ResNN.transferFix8FromN7V20P50.ResNN.transferFromN7V20P50.min}\%.  
This is reasonable since $\textrm{N7}_a$ and $\textrm{N7}_b$ have the same design rules and illumination shapes, 
and the only difference lies in the resist materials. 
Therefore, the feature extraction layers are supposed to remain almost the same. 
With the sizes of the training dataset increasing to 15\% and 20\%, the differences in the accuracy become smaller, 
because there are enough data to find good configurations for the networks. 
Since knowledge transfer is remarkably effective with ResNet $\textrm{TF}_8$, we do not see the room for further improvement with active learning. 
Thus we did not plot the curves for that. 

\subsection{Impact of Various Source Domains}
\label{sec:ImpactofVariousSourceDomains}

\begin{figure}[tb!]
\vspace{-.1in}
\centering
\includegraphics[]{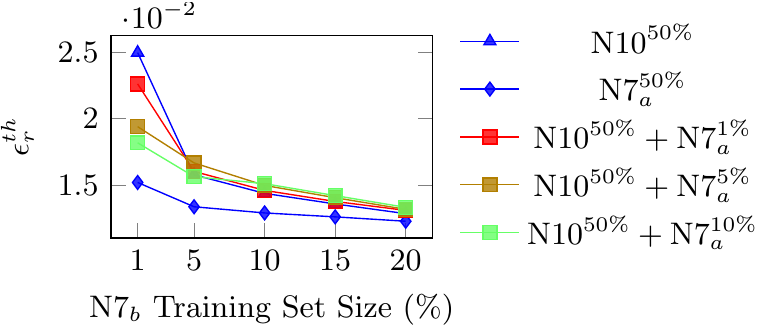}
\caption{~Testing accuracy of ResNet $\textrm{TF}_0$ for $\textrm{N7}_b$ from different source domain datasets.}
\label{fig:N7FromDiffSources}
\end{figure}

\begin{figure}[tb]
\centering
\includegraphics[]{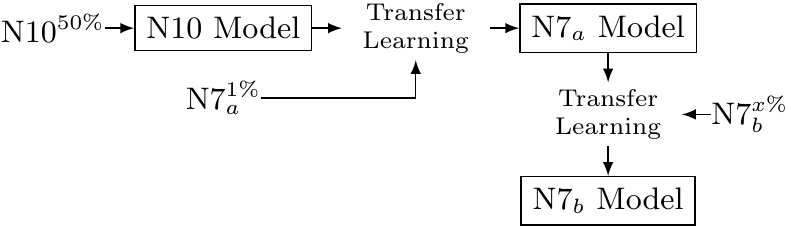}
\caption{~Transfer learning from 50\% of N10 dataset and 1\% of $\textrm{N7}_a$ dataset (i.e., $\textrm{N10}^{50\%}+\textrm{N7}_a^{1\%}$) to $\textrm{N7}_b$ with $x\%$ of $\textrm{N7}_b$ dataset.}
\label{fig:TransferTwice}
\end{figure}

In transfer learning, the correlation between the datasets of source and target domains is critical to the effectiveness of knowledge transfer. 
Thus, we explore the impacts of source domain datasets on the accuracy of modeling for the target domain. 
Figure~\ref{fig:N7FromDiffSources} plots the testing errors of learning $\textrm{N7}_b$ using ResNet $\textrm{TF}_0$ with various source domain datasets. 
Curves ``$\textrm{N10}^{50\%}$'' and ``$\textrm{N7}_a^{50\%}$'' indicate that 50\% of the N10 or the $\textrm{N7}_a$ dataset is used to train source domain models, respectively. 
Curve ``$\textrm{N10}^{50\%}+\textrm{N7}_a^{1\%}$'' describes the situation where we have 50\% of the N10 dataset and 1\% of the $\textrm{N7}_a$ dataset for training. 
In this case, as shown in Figure~\ref{fig:TransferTwice}, we first use the 50\% N10 data to train the first source domain model;
then train the second source domain model using the first model as the starting point with the 1\% $\textrm{N7}_a$ data; 
in the end, the target domain model for $\textrm{N7}_b$ is trained using the second model as the starting point with $\textrm{N7}_b$ data. 
Curves ``$\textrm{N10}^{50\%}+\textrm{N7}_a^{5\%}$'' and ``$\textrm{N10}^{50\%}+\textrm{N7}_a^{10\%}$'' are similar, simply with different amounts of $\textrm{N7}_a$ data for training. 

The knowledge from $\textrm{N7}_a^{50\%}$ is the most effective for $\textrm{N7}_b$ due to the minor difference in resist materials between two datasets. 
For the rest curves, 
the accuracy of $\textrm{N10}^{50\%}+\textrm{N7}_a^{5\%}$ and $\textrm{N10}^{50\%}+\textrm{N7}_a^{10\%}$ is in general better than or at least comparable to that of $\textrm{N10}^{50\%}$. 
This indicates that having more data from closer datasets to the target dataset, e.g., $\textrm{N7}_a$, is still helpful. 

\subsection{Improvement in Data Efficiency}
\label{sec:ImprovementInDataEfficiency}

Table~\ref{tab:DataEfficiency} presents the accuracy metrics, i.e., relative threshold RMS error ($\epsilon_r^{th}$) and CD RMS error ($\epsilon^{CD}$), for learning $\textrm{N7}_b$ from various source domain datasets. 
Since we consider the data efficiency of different learning schemes, we focus on the small training dataset for $\textrm{N7}_b$, from 1\% to 20\%. 
Situations such as no source domain data ($\emptyset$), 
only source domain data from N10 ($\textrm{N10}^{50\%}$), 
only source domain data from $\textrm{N7}_a$ ($\textrm{N7}_a^{50\%}$), and combined source domain datasets, are examined. 
As mentioned in Section~\ref{sec:Preliminary}, the fidelity between relative threshold RMS error and CD RMS error is very consistent, so they share almost the same trends. 
Transfer learning with any source domain dataset enables an average improvement of \DeltaScale{\UseMacro{N7ToN7Final.CNN.N7.V23.ratio.CDPredictRMS}}{\UseMacro{N10ToN7Final.CNN.transferFromN10P50.N7.V23.ratio.CDPredictRMS}}{100}\% to \DeltaScale{\UseMacro{N7ToN7Final.CNN.N7.V23.ratio.CDPredictRMS}}{\UseMacro{N7ToN7Final.ResNN.transferFromN7V20P50.N7.V23.ratio.CDPredictRMS}}{100}\% from that without knowledge transfer. 
In small training datasets of $\textrm{N7}_b$, ResNet also achieves around \DeltaScale{\UseMacro{N10ToN7Final.CNN.transferFromN10P50.N7.V23.ratio.CDPredictRMS}}{\UseMacro{N10ToN7Final.ResNN.transferFromN10P50.N7.V23.ratio.CDPredictRMS}}{100}\% better performance on average than CNN in the transfer learning scheme. 
Enabling active learning together with transfer learning allows additional \DeltaScale{\UseMacro{N10ToN7Final.ResNN.transferFromN10P50.N7.V23.ratio.CDPredictRMS}}{\UseMacro{ActiveLearning.KMeans.ResNN.transferFromN10P50.N7.V23.ratio.CDPredictRMS}}{100}\% accuracy improvement on average compared with transfer learning only for ResNet. 
At 1\% of $\textrm{N7}_b$, combined source domain datasets have better performance compared with $\textrm{N10}^{50\%}$ only, 
but the benefits vanish with the increase of the $\textrm{N7}_b$ dataset. 

\begin{figure}
    \centering
    \hspace{.2in}
    \begin{tikzpicture}
        \node[rectangle, draw, fill=black, minimum width=0.3cm, minimum height=0.2cm, inner sep=0pt, outer sep=0pt](CNNImage){};
        \node[right=0.1cm of CNNImage, inner sep=0pt, outer sep=0pt](CNNText){CNN}; 
        \node[right=0.2cm of CNNText, rectangle, draw, minimum width=0.3cm, minimum height=0.2cm, fill=orange, fill opacity=0.8, inner sep=0pt, outer sep=0pt](CNNTF0Image){};
        \node[right=0.1cm of CNNTF0Image, inner sep=0pt, outer sep=0pt](CNNTF0Text){CNN $\textrm{TF}_0$};
        \node[right=0.2cm of CNNTF0Text, rectangle, draw, minimum width=0.3cm, minimum height=0.2cm, fill=blue, fill opacity=0.8, inner sep=0pt, outer sep=0pt](ResNetTF0Image){};
        \node[right=0.1cm of ResNetTF0Image, inner sep=0pt, outer sep=0pt](ResNetTF0Text){ResNet $\textrm{TF}_0$};
        \node[right=0.2cm of ResNetTF0Text, rectangle, draw, minimum width=0.3cm, minimum height=0.2cm, fill=sky, fill opacity=0.8, inner sep=0pt, outer sep=0pt](ResNetTF0ALImage){};
        \node[right=0.1cm of ResNetTF0ALImage, inner sep=0pt, outer sep=0pt](ResNetTF0ALText){ResNet $\textrm{TF}_0$+AL\footnotemark[\value{footnote}]};
    \end{tikzpicture}
    \subfloat[]{\includegraphics[]{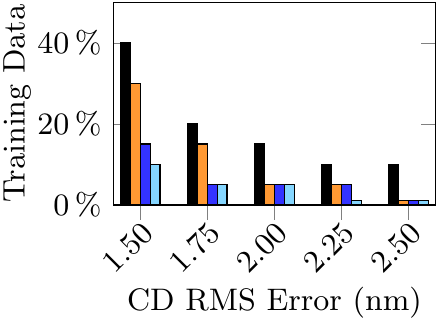}\label{fig:N7V23FromN10DataRequired}}\hfill%
    \subfloat[]{\includegraphics[]{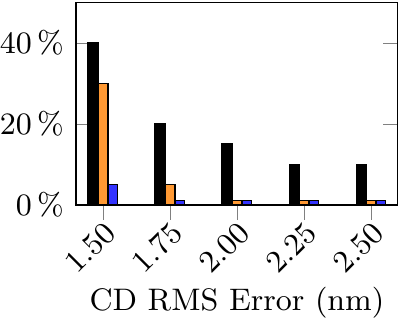}\label{fig:N7V23FromN7DataRequired}}
    \caption{~Amount of training data required for $\textrm{N7}_b$ given target CD RMS errors when \protect\subref{fig:N7V23FromN10DataRequired} 50\% N10 dataset is available 
        or \protect\subref{fig:N7V23FromN7DataRequired} 50\% $\textrm{N7}_a$ dataset is available.
    }
    \label{fig:N7V23DataRequired}
\end{figure}

In real manufacturing, models are usually calibrated to satisfy a target accuracy or target CD RMS error. 
Figure~\ref{fig:N7V23DataRequired} demonstrates the amount of training data required in the target domain for learning the $\textrm{N7}_b$ model. 
Curve ``CNN'' does not involve any knowledge transfer, while curves ``CNN $\textrm{TF}_0$'' and ``ResNet $\textrm{TF}_0$'' utilize transfer learning in CNN and ResNet, respectively. 
The curves in Figure~\subref*{fig:N7V23FromN10DataRequired} assume the availability of N10 data. 
Consider the CD RMS error from 1.5$nm$ to 2.5$nm$, which is around 10\% of the half pitch for N7 contacts. 
This range of accuracy is also comparable to that of the state-of-the-art CNN \cite{RESIST_SPIE2017_Watanabe}.  
ResNet $\textrm{TF}_0$ requires significantly fewer data than both CNN and CNN $\textrm{TF}_0$. 
For instance, when the target CD error is 1.75$nm$, ResNet $\textrm{TF}_0$ demands \UseMacro{ResNetFromN10P50.targetCD.1.75}\% training data from $\textrm{N7}_b$, while CNN requires \UseMacro{CNN.targetCD.1.75}\% and CNN $\textrm{TF}_0$ requires \UseMacro{CNNFromN10P50.targetCD.1.75}\%.  
By enabling active learning, ResNet $\textrm{TF}_0$+AL further reduces data requirement from ResNet $\textrm{TF}_0$, 
e.g., \RatioScale{\UseMacro{ResNetFromN10P50.targetCD.1.5}}{\UseMacro{KMeansResNetFromN10P50.targetCD.1.5}}{1}X 
and \RatioScale{\UseMacro{CNN.targetCD.1.5}}{\UseMacro{KMeansResNetFromN10P50.targetCD.1.5}}{1}X
fewer training data than ResNet $\textrm{TF}_0$ and CNN for 1.5$nm$, respectively. 
Figure~\subref*{fig:N7V23FromN7DataRequired} considers the transfer from $\textrm{N7}_a$ to $\textrm{N7}_b$. 
Both ResNet $\textrm{TF}_0$ and CNN $\textrm{TF}_0$ only require \UseMacro{ResNetFromN7P50.targetCD.2}\% training data from $\textrm{N7}_b$ for most target CD RMS errors, 
where CNN $\textrm{TF}_0$ cannot achieve the accuracy unless given \UseMacro{CNNFromN7P50.targetCD.1.5}\% data. 
Overall, ResNet $\textrm{TF}_0$ can achieve \UseMacro{TCADMinDataReduction}-\UseMacro{TCADMaxDataReduction}X reduction of training data within this range compared with CNN. 
It needs to mention that 1\% of dataset only correspond to fewer than 40 samples owing to the data augmentation, 
indicating only thresholds of 40 clips are required.  

\begin{table*}[tb]
\centering
\caption{Relative Threshold RMS Error and CD RMS Error for $\textrm{N7}_b$ with Different Source Domain Datasets}
\label{tab:DataEfficiency}
\resizebox{\textwidth}{!} 
{
\begin{tabular}{|c|c|c|c|c|c|c|c|c|c|c|c|c|c|c|c|c|c|}
\hline
\multicolumn{2}{|c|}{\begin{tabular}[c]{@{}c@{}}Source\\ Datasets\end{tabular}}  
    & \multicolumn{2}{c|}{$\emptyset$}                                                                    
    & \multicolumn{6}{c|}{$\textrm{N10}^{50\%}$}                                                                                                                                        
    & \multicolumn{4}{c|}{$\textrm{N7}_a^{50\%}$}                                                                                                                                           
    & \multicolumn{2}{c|}{\begin{tabular}[c]{@{}c@{}}$\textrm{N10}^{50\%}+\textrm{N7}_a^{5\%}$\end{tabular}} & \multicolumn{2}{c|}{\begin{tabular}[c]{@{}c@{}}$\textrm{N10}^{50\%}+\textrm{N7}_a^{10\%}$\end{tabular}} \\ \hline
\multicolumn{2}{|c|}{\begin{tabular}[c]{@{}c@{}}Neural\\ Networks\end{tabular}} 
& \multicolumn{2}{c|}{CNN}                                                                  
& \multicolumn{2}{c|}{\begin{tabular}[c]{@{}c@{}}CNN $\textrm{TF}_0$\end{tabular}}    
& \multicolumn{2}{c|}{\begin{tabular}[c]{@{}c@{}}ResNet $\textrm{TF}_0$\end{tabular}}     
& \multicolumn{2}{c|}{\begin{tabular}[c]{@{}c@{}}ResNet $\textrm{TF}_0$+AL\footnotemark[\value{footnote}]\end{tabular}}     
& \multicolumn{2}{c|}{\begin{tabular}[c]{@{}c@{}}CNN $\textrm{TF}_0$\end{tabular}}        
& \multicolumn{2}{c|}{\begin{tabular}[c]{@{}c@{}}ResNet $\textrm{TF}_0$\end{tabular}}     
& \multicolumn{2}{c|}{\begin{tabular}[c]{@{}c@{}}ResNet $\textrm{TF}_0$\end{tabular}}                      
& \multicolumn{2}{c|}{\begin{tabular}[c]{@{}c@{}}ResNet $\textrm{TF}_0$\end{tabular}}                       \\ \hline
\multicolumn{2}{|c|}{}                                                          
& \begin{tabular}[c]{@{}c@{}}$\epsilon_r^{th}$\\ ($10^{-2}$)\end{tabular} 
& $\epsilon^{CD}$ 
& \begin{tabular}[c]{@{}c@{}}$\epsilon_r^{th}$\\ ($10^{-2}$)\end{tabular} 
& $\epsilon^{CD}$ 
& \begin{tabular}[c]{@{}c@{}}$\epsilon_r^{th}$\\ ($10^{-2}$)\end{tabular} 
& $\epsilon^{CD}$ 
& \begin{tabular}[c]{@{}c@{}}$\epsilon_r^{th}$\\ ($10^{-2}$)\end{tabular} 
& $\epsilon^{CD}$ 
& \begin{tabular}[c]{@{}c@{}}$\epsilon_r^{th}$\\ ($10^{-2}$)\end{tabular} 
& $\epsilon^{CD}$ 
& \begin{tabular}[c]{@{}c@{}}$\epsilon_r^{th}$\\ ($10^{-2}$)\end{tabular} 
& $\epsilon^{CD}$ 
& \begin{tabular}[c]{@{}c@{}}$\epsilon_r^{th}$\\ ($10^{-2}$)\end{tabular}          
& $\epsilon^{CD}$         
& \begin{tabular}[c]{@{}c@{}}$\epsilon_r^{th}$\\ ($10^{-2}$)\end{tabular}          
& $\epsilon^{CD}$          \\ \hline
\multirow{5}{*}{$\textrm{N7}_b$}                      & 1\%                     &\UseMacro{N7ToN7Final.CNN.N7.V23.0.01.thresholdPredictRelativeRMS} &\UseMacro{N7ToN7Final.CNN.N7.V23.0.01.CDPredictRMS} &\UseMacro{N10ToN7Final.CNN.transferFromN10P50.N7.V23.0.01.thresholdPredictRelativeRMS} &\UseMacro{N10ToN7Final.CNN.transferFromN10P50.N7.V23.0.01.CDPredictRMS} &\UseMacro{N10ToN7Final.ResNN.transferFromN10P50.N7.V23.0.01.thresholdPredictRelativeRMS} &\UseMacro{N10ToN7Final.ResNN.transferFromN10P50.N7.V23.0.01.CDPredictRMS}  &\UseMacro{ActiveLearning.KMeans.ResNN.transferFromN10P50.N7.V23.0.01.thresholdPredictRelativeRMS} &\UseMacro{ActiveLearning.KMeans.ResNN.transferFromN10P50.N7.V23.0.01.CDPredictRMS} &\UseMacro{N7ToN7Final.CNN.transferFromN7V20P50.N7.V23.0.01.thresholdPredictRelativeRMS} &\UseMacro{N7ToN7Final.CNN.transferFromN7V20P50.N7.V23.0.01.CDPredictRMS} &\UseMacro{N7ToN7Final.ResNN.transferFromN7V20P50.N7.V23.0.01.thresholdPredictRelativeRMS} &\UseMacro{N7ToN7Final.ResNN.transferFromN7V20P50.N7.V23.0.01.CDPredictRMS} &\UseMacro{N7ToN7Final.ResNN.transferFromN7V20P5.N7.V23.0.01.thresholdPredictRelativeRMS} &\UseMacro{N7ToN7Final.ResNN.transferFromN7V20P5.N7.V23.0.01.CDPredictRMS} &\UseMacro{N7ToN7Final.ResNN.transferFromN7V20P10.N7.V23.0.01.thresholdPredictRelativeRMS} &\UseMacro{N7ToN7Final.ResNN.transferFromN7V20P10.N7.V23.0.01.CDPredictRMS} \\ \cline{2-18} 
                                                      & 5\%                     &\UseMacro{N7ToN7Final.CNN.N7.V23.0.05.thresholdPredictRelativeRMS} &\UseMacro{N7ToN7Final.CNN.N7.V23.0.05.CDPredictRMS} &\UseMacro{N10ToN7Final.CNN.transferFromN10P50.N7.V23.0.05.thresholdPredictRelativeRMS} &\UseMacro{N10ToN7Final.CNN.transferFromN10P50.N7.V23.0.05.CDPredictRMS} &\UseMacro{N10ToN7Final.ResNN.transferFromN10P50.N7.V23.0.05.thresholdPredictRelativeRMS} &\UseMacro{N10ToN7Final.ResNN.transferFromN10P50.N7.V23.0.05.CDPredictRMS} &\UseMacro{ActiveLearning.KMeans.ResNN.transferFromN10P50.N7.V23.0.05.thresholdPredictRelativeRMS} &\UseMacro{ActiveLearning.KMeans.ResNN.transferFromN10P50.N7.V23.0.05.CDPredictRMS} &\UseMacro{N7ToN7Final.CNN.transferFromN7V20P50.N7.V23.0.05.thresholdPredictRelativeRMS} &\UseMacro{N7ToN7Final.CNN.transferFromN7V20P50.N7.V23.0.05.CDPredictRMS} &\UseMacro{N7ToN7Final.ResNN.transferFromN7V20P50.N7.V23.0.05.thresholdPredictRelativeRMS} &\UseMacro{N7ToN7Final.ResNN.transferFromN7V20P50.N7.V23.0.05.CDPredictRMS} &\UseMacro{N7ToN7Final.ResNN.transferFromN7V20P5.N7.V23.0.05.thresholdPredictRelativeRMS} &\UseMacro{N7ToN7Final.ResNN.transferFromN7V20P5.N7.V23.0.05.CDPredictRMS} &\UseMacro{N7ToN7Final.ResNN.transferFromN7V20P10.N7.V23.0.05.thresholdPredictRelativeRMS} &\UseMacro{N7ToN7Final.ResNN.transferFromN7V20P10.N7.V23.0.05.CDPredictRMS} \\ \cline{2-18} 
                                                      & 10\%                    &\UseMacro{N7ToN7Final.CNN.N7.V23.0.1.thresholdPredictRelativeRMS}  &\UseMacro{N7ToN7Final.CNN.N7.V23.0.1.CDPredictRMS}  &\UseMacro{N10ToN7Final.CNN.transferFromN10P50.N7.V23.0.1.thresholdPredictRelativeRMS}  &\UseMacro{N10ToN7Final.CNN.transferFromN10P50.N7.V23.0.1.CDPredictRMS}  &\UseMacro{N10ToN7Final.ResNN.transferFromN10P50.N7.V23.0.1.thresholdPredictRelativeRMS}  &\UseMacro{N10ToN7Final.ResNN.transferFromN10P50.N7.V23.0.1.CDPredictRMS}  &\UseMacro{ActiveLearning.KMeans.ResNN.transferFromN10P50.N7.V23.0.1.thresholdPredictRelativeRMS} &\UseMacro{ActiveLearning.KMeans.ResNN.transferFromN10P50.N7.V23.0.1.CDPredictRMS} &\UseMacro{N7ToN7Final.CNN.transferFromN7V20P50.N7.V23.0.1.thresholdPredictRelativeRMS}  &\UseMacro{N7ToN7Final.CNN.transferFromN7V20P50.N7.V23.0.1.CDPredictRMS}  &\UseMacro{N7ToN7Final.ResNN.transferFromN7V20P50.N7.V23.0.1.thresholdPredictRelativeRMS}  &\UseMacro{N7ToN7Final.ResNN.transferFromN7V20P50.N7.V23.0.1.CDPredictRMS}  &\UseMacro{N7ToN7Final.ResNN.transferFromN7V20P5.N7.V23.0.1.thresholdPredictRelativeRMS}  &\UseMacro{N7ToN7Final.ResNN.transferFromN7V20P5.N7.V23.0.1.CDPredictRMS}  &\UseMacro{N7ToN7Final.ResNN.transferFromN7V20P10.N7.V23.0.1.thresholdPredictRelativeRMS}  &\UseMacro{N7ToN7Final.ResNN.transferFromN7V20P10.N7.V23.0.1.CDPredictRMS}  \\ \cline{2-18} 
                                                      & 15\%                    &\UseMacro{N7ToN7Final.CNN.N7.V23.0.15.thresholdPredictRelativeRMS} &\UseMacro{N7ToN7Final.CNN.N7.V23.0.15.CDPredictRMS} &\UseMacro{N10ToN7Final.CNN.transferFromN10P50.N7.V23.0.15.thresholdPredictRelativeRMS} &\UseMacro{N10ToN7Final.CNN.transferFromN10P50.N7.V23.0.15.CDPredictRMS} &\UseMacro{N10ToN7Final.ResNN.transferFromN10P50.N7.V23.0.15.thresholdPredictRelativeRMS} &\UseMacro{N10ToN7Final.ResNN.transferFromN10P50.N7.V23.0.15.CDPredictRMS} &\UseMacro{ActiveLearning.KMeans.ResNN.transferFromN10P50.N7.V23.0.15.thresholdPredictRelativeRMS} &\UseMacro{ActiveLearning.KMeans.ResNN.transferFromN10P50.N7.V23.0.15.CDPredictRMS} &\UseMacro{N7ToN7Final.CNN.transferFromN7V20P50.N7.V23.0.15.thresholdPredictRelativeRMS} &\UseMacro{N7ToN7Final.CNN.transferFromN7V20P50.N7.V23.0.15.CDPredictRMS} &\UseMacro{N7ToN7Final.ResNN.transferFromN7V20P50.N7.V23.0.15.thresholdPredictRelativeRMS} &\UseMacro{N7ToN7Final.ResNN.transferFromN7V20P50.N7.V23.0.15.CDPredictRMS} &\UseMacro{N7ToN7Final.ResNN.transferFromN7V20P5.N7.V23.0.15.thresholdPredictRelativeRMS} &\UseMacro{N7ToN7Final.ResNN.transferFromN7V20P5.N7.V23.0.15.CDPredictRMS} &\UseMacro{N7ToN7Final.ResNN.transferFromN7V20P10.N7.V23.0.15.thresholdPredictRelativeRMS} &\UseMacro{N7ToN7Final.ResNN.transferFromN7V20P10.N7.V23.0.15.CDPredictRMS} \\ \cline{2-18} 
                                                      & 20\%                    &\UseMacro{N7ToN7Final.CNN.N7.V23.0.2.thresholdPredictRelativeRMS}  &\UseMacro{N7ToN7Final.CNN.N7.V23.0.2.CDPredictRMS}  &\UseMacro{N10ToN7Final.CNN.transferFromN10P50.N7.V23.0.2.thresholdPredictRelativeRMS}  &\UseMacro{N10ToN7Final.CNN.transferFromN10P50.N7.V23.0.2.CDPredictRMS}  &\UseMacro{N10ToN7Final.ResNN.transferFromN10P50.N7.V23.0.2.thresholdPredictRelativeRMS}  &\UseMacro{N10ToN7Final.ResNN.transferFromN10P50.N7.V23.0.2.CDPredictRMS}  &\UseMacro{ActiveLearning.KMeans.ResNN.transferFromN10P50.N7.V23.0.2.thresholdPredictRelativeRMS} &\UseMacro{ActiveLearning.KMeans.ResNN.transferFromN10P50.N7.V23.0.2.CDPredictRMS} &\UseMacro{N7ToN7Final.CNN.transferFromN7V20P50.N7.V23.0.2.thresholdPredictRelativeRMS}  &\UseMacro{N7ToN7Final.CNN.transferFromN7V20P50.N7.V23.0.2.CDPredictRMS}  &\UseMacro{N7ToN7Final.ResNN.transferFromN7V20P50.N7.V23.0.2.thresholdPredictRelativeRMS}  &\UseMacro{N7ToN7Final.ResNN.transferFromN7V20P50.N7.V23.0.2.CDPredictRMS}  &\UseMacro{N7ToN7Final.ResNN.transferFromN7V20P5.N7.V23.0.2.thresholdPredictRelativeRMS}  &\UseMacro{N7ToN7Final.ResNN.transferFromN7V20P5.N7.V23.0.2.CDPredictRMS}  &\UseMacro{N7ToN7Final.ResNN.transferFromN7V20P10.N7.V23.0.2.thresholdPredictRelativeRMS}  &\UseMacro{N7ToN7Final.ResNN.transferFromN7V20P10.N7.V23.0.2.CDPredictRMS}  \\ \hline\hline
\multicolumn{2}{|c|}{ratio}                                                     &\UseMacro{N7ToN7Final.CNN.N7.V23.ratio.thresholdPredictRelativeRMS}&\UseMacro{N7ToN7Final.CNN.N7.V23.ratio.CDPredictRMS}&\UseMacro{N10ToN7Final.CNN.transferFromN10P50.N7.V23.ratio.thresholdPredictRelativeRMS}&\UseMacro{N10ToN7Final.CNN.transferFromN10P50.N7.V23.ratio.CDPredictRMS}&\UseMacro{N10ToN7Final.ResNN.transferFromN10P50.N7.V23.ratio.thresholdPredictRelativeRMS}&\UseMacro{N10ToN7Final.ResNN.transferFromN10P50.N7.V23.ratio.CDPredictRMS}&\UseMacro{ActiveLearning.KMeans.ResNN.transferFromN10P50.N7.V23.ratio.thresholdPredictRelativeRMS}&\UseMacro{ActiveLearning.KMeans.ResNN.transferFromN10P50.N7.V23.ratio.CDPredictRMS}&\UseMacro{N7ToN7Final.CNN.transferFromN7V20P50.N7.V23.ratio.thresholdPredictRelativeRMS}&\UseMacro{N7ToN7Final.CNN.transferFromN7V20P50.N7.V23.ratio.CDPredictRMS}&\UseMacro{N7ToN7Final.ResNN.transferFromN7V20P50.N7.V23.ratio.thresholdPredictRelativeRMS}&\UseMacro{N7ToN7Final.ResNN.transferFromN7V20P50.N7.V23.ratio.CDPredictRMS}&\UseMacro{N7ToN7Final.ResNN.transferFromN7V20P5.N7.V23.ratio.thresholdPredictRelativeRMS}&\UseMacro{N7ToN7Final.ResNN.transferFromN7V20P5.N7.V23.ratio.CDPredictRMS}&\UseMacro{N7ToN7Final.ResNN.transferFromN7V20P10.N7.V23.ratio.thresholdPredictRelativeRMS}&\UseMacro{N7ToN7Final.ResNN.transferFromN7V20P10.N7.V23.ratio.CDPredictRMS}\\ \hline
\end{tabular}
}
\end{table*}

%% file: conclu.tex
\section{Conclusion}
\label{sec:Conclusion}

A transfer learning framework with a clustering based active data selection on residual neural networks is proposed for resist modeling. 
The combination of transfer learning and active learning for ResNet is able to achieve high accuracy with very few data from the target domain, 
under various situations for knowledge transfer, indicating high data efficiency.  
Extensive experiments demonstrate that the proposed techniques can achieve \UseMacro{TCADMinDataReduction}-\UseMacro{TCADMaxDataReduction}X reduction according to various requirements of accuracy comparable to the state-of-the-art learning approach. 
It is shown that the performance of transfer learning differs from dataset to dataset and is worth exploring to see the correlation between datasets. 
Active selection of data samples is also useful to guide the generation of mask designs for model calibration in manufacturing. 
Examining the quantitative relation between the correlation of datasets and performance of transfer learning is valuable in the future. 
\textcolor{black}{
There is still room to improve the effectiveness of knowledge transfer from N10 to N7 datasets. 
Therefore, in the future, we will actively explore other learning techniques to further improve the accuracy, 
such as preprocessing steps like scaling and normalization, various regularization techniques, and semi-supervised learning. 
}

\section*{Acknowledge}

This project is supported in part by Toshiba Memory Corporation, NSF, and the University Graduate Continuing Fellowship from the University of Texas at Austin. 
The authors would like to thank Memory Lithography Group from Toshiba Memory Corporation 
and Dr. Kai Zhong from the Computer Science Department of UT Austin for helpful discussions and feedback.

%% file: appdx.tex
\onecolumn
\appendix
\label{sec:appdx}


\DefMacro{Inequality1}{a}
\DefMacro{Inequality2}{b}
\DefMacro{Inequality3}{c}
\DefMacro{Inequality4}{d}
\DefMacro{Inequality5}{e}
\DefMacro{Inequality6}{f}
\DefMacro{Inequality7}{g}
\DefMacro{Inequality8}{h}
\DefMacro{Inequality9}{i}


\noindent 
Proof of Lemma~\ref{lemma:CNNLipschitz}. 
\begin{proof}
    We assume the output after a series of convolution and fully connected layers is the prediction of the CNN for regression. 
    Consider two inputs $\bm{x}$ and $\bm{x}'$, with their representation $\bm{x}^{(d)}$ and $\bm{x}'^{(d)}$. 
    We first show the Lipschitz property of convolution layers and fully connected layers. 
    Any convolution or fully connected layer can be denoted as $\bm{x}^{(d)}_j = \sum_{i} w^{(d)}_{i, j} \bm{x}^{(d-1)}_i$. 
    By assuming $\sum_{i} |w^{(d)}_{i, j}| \le \alpha, \forall i, j, d$, we can state, 
    \begin{equation}
        \norm{\bm{x}^{(d)}_j - \bm{x}'^{(d)}_j} \le \alpha \norm{\bm{x}^{(d-1)} - \bm{x}'^{(d-1)}}. 
    \end{equation}
    Let $n^{(d)}$ be the dimension of $\bm{x}^{(d)}$, which is bounded by $N$, 
    \begin{equation}
    \begin{split}
        \norm{\bm{x}^{(d)} - \bm{x}'^{(d)}} & = \sqrt{\sum_j \norm{\bm{x}^{(d)}_j - \bm{x}'^{(d)}_j}^2}, \\
                                            & \le \sqrt{n^{(d)} \alpha^2 \norm{\bm{x}^{(d-1)} - \bm{x}'^{(d-1)}}^2}, \\
                                            & \le \alpha \sqrt{n^{(d)}} \norm{\bm{x}^{(d-1)} - \bm{x}'^{(d-1)}}, \\ 
                                            & \le \alpha \sqrt{N} \norm{\bm{x}^{(d-1)} - \bm{x}'^{(d-1)}}. 
    \end{split}
    \end{equation}

    We then consider ReLU and max-pooling layers. 
    For any ReLU layer, it is straightforward to verify the following inequality, 
    \begin{equation}
        |\max(0, a) - \max(0, b)| \le |a-b|. 
    \end{equation}
    Any max-pooling layer can be viewed as a convolution layer in which only one weight is 1 and others are 0. 
    Thus, we can state for ReLU and max-pooling layers, 
    \begin{equation}
        \norm{\bm{x}^{(d)} - \bm{x}'^{(d)}} \le \norm{\bm{x}^{(d-1)} - \bm{x}'^{(d-1)}}. 
    \end{equation}

    Combining the Lipschitz property of all layers of CNN, 
    \begin{equation}
        \norm{CNN(\bm{x}; \bm{w}) - CNN(\bm{x}'; \bm{w})} \le (\alpha\sqrt{N})^{n_c+n_{fc}} \norm{\bm{x} - \bm{x}'},  
        \label{eq:CNNLipschitzBound}
    \end{equation}
    where $\bm{w}$ is the weights for CNN. 

    For ResNet, a shortcut connection can be viewed as a layer $d$ which takes input from layer $d-1$ and layer $d'$, i.e., $\bm{x}^{(d)} = \bm{x}^{(d-1)} + \bm{x}^{(d')}$, where $d-1 > d'$. 
    Then we can state, 
    \begin{equation}
    \begin{split}
        \norm{\bm{x}^{(d)} - \bm{x}^{(d)}} & = \norm{\bm{x}^{(d-1)} + \bm{x}^{(d')} - \bm{x}'^{(d-1)} - \bm{x}'^{(d')}}, \\
                                           & \le \norm{\bm{x}^{(d-1)} - \bm{x}'^{(d-1)} } + \norm{\bm{x}^{(d')} - \bm{x}'^{(d')}}, \\
                                           & \le \big( (\alpha \sqrt{N})^{d-d'-1}+1 \big) \norm{\bm{x}^{(d')} - \bm{x}'^{(d')}}, \\
                                           & \le (1+\alpha \sqrt{N})^{d-d'-1} \norm{\bm{x}^{(d')} - \bm{x}'^{(d')}}.  
    \end{split}
    \end{equation}
    Therefore, combining all layers of ResNet, 
    \begin{equation}
        \norm{ResNet(\bm{x}; \bm{w}) - ResNet(\bm{x}'; \bm{w})} \le (1+\alpha\sqrt{N})^{n_c+n_{fc}} \norm{\bm{x} - \bm{x}'}.  
        \label{eq:ResNetLipschitzBound}
    \end{equation}

    We combine Eq.~\eqref{eq:CNNLipschitzBound} and Eq.~\eqref{eq:ResNetLipschitzBound} for generalization to both CNN and ResNet. 

\end{proof}


\noindent 
Proof of Lemma~\ref{lemma:LossLipschitz}. 
{\color{black}
\begin{proof}
    We first bound $\abs{ \hat{f}(\bm{x}_i) }$, 
    \begin{equation}
    \begin{split}
        \abs{ \hat{f}(\bm{x}_i) } & \le \abs{\hat{f}(\bm{x}_0)} + \abs{ \hat{f}(\bm{x}_i) - \hat{f}(\bm{x}_0) }, \\
                                  & \le \abs{y_0} + \abs{\delta} + \lambda^{\hat{f}} \norm{ \bm{x}_i - \bm{x}_0 }, \\
                                  & \le b_2 + \abs{\delta} + \lambda^{\hat{f}} (\norm{ \bm{x}_i } + \norm{ \bm{x}_0 }), \\
                                  & \le b_2 + \abs{\delta} + 2 \lambda^{\hat{f}} b_1. \\
    \end{split}
    \end{equation}
    Then prove the Lipschitz-continuity of square loss function, 
    \begin{equation}
    \begin{split}
        \abs { l_{\bm{w}_{\bm{s}}}(\bm{x}_i, y_i) - l_{\bm{w}_{\bm{s}}}(\bm{x}_j, y_j) } & = \abs { (y_i - \hat{f}(\bm{x}_i))^2 -  (y_j - \hat{f}(\bm{x}_j))^2 }, \\
                                                        & \le \abs { y_i - y_j - \hat{f}(\bm{x}_i) + \hat{f}(\bm{x}_j) } \abs { y_i + y_j - \hat{f}(\bm{x}_i) - \hat{f}(\bm{x}_j) }, \\
                                                        & \le (\abs{y_i - y_j} + \abs{\hat{f}(\bm{x}_i) - \hat{f}(\bm{x}_j)}) 
        (\abs{ y_i } + \abs{ y_j } + \abs{ \hat{f}(\bm{x}_i) } + \abs { \hat{f}(\bm{x}_j) }), \\
        & \le (\abs{y_i - y_j} + \lambda^{\hat{f}}\norm{\bm{x}_i - \bm{x}_j}) 
        ( 4 b_2 + 2 \abs{\delta} + 4 \lambda^{\hat{f}} b_1 ), \\
        & \le ( 4 \lambda^{\hat{f}} b_1 + 4 b_2  + 2 \abs{\delta} ) \cdot \max(1, \lambda^{\hat{f}}) (\abs{y_i - y_j} + \norm{\bm{x}_i - \bm{x}_j}).  
    \end{split}
    \end{equation}
\end{proof}
}


\noindent 
Proof of Theorem~\ref{theorem:AvgLossBound}. 
{\color{black}
\begin{proof}

\begin{equation}
\begin{split}
    \abs{l_{\bm{w}_{\bm{s}}}(\bm{x}_i, y_i) - l_{\bm{w}_{\bm{s}}}(\bm{x}_j, y_j)} 
    & \stackrel{(\UseMacro{Inequality1})}{\le} \lambda^l (\abs{y_i - y_j} + \norm{ \bm{x}_i - \bm{x}_j }). 
    \label{eq:ResistModel:yhatLipschitz}
\end{split}
\end{equation}
Inequality (\UseMacro{Inequality1}) uses the Lipschitz property of the loss function. 

\begin{equation}
\begin{split}
    \abs{y_i - y_j}
    & = \abs{f(\bm{x}_i) + \epsilon_i - f(\bm{x}_j) - \epsilon_j}, \\
    & \le \abs{f(\bm{x}_i) - f(\bm{x}_j)} + \abs{\epsilon_i} + \abs{\epsilon_j}, \\
    & \stackrel{(\UseMacro{Inequality2})}{\le} \lambda^f \norm{ \bm{x}_i - \bm{x}_j } + \abs{\epsilon_i} + \abs{\epsilon_j}, 
    \label{eq:ResistModel:yLipschitz}
\end{split}
\end{equation}
Inequality (\UseMacro{Inequality2}) uses the Lipschitz property of the ground truth function $f$.

Combine previous two inequalities Eq.~\eqref{eq:ResistModel:yhatLipschitz} and Eq.~\eqref{eq:ResistModel:yLipschitz}, we have, 
\begin{equation}
\begin{split}
    \abs{l_{\bm{w}_{\bm{s}}}(\bm{x}_i, y_i) - l_{\bm{w}_{\bm{s}}}(\bm{x}_j, y_j)} 
    & \le \lambda^l (\lambda^f \norm{ \bm{x}_i - \bm{x}_j } + \abs{\epsilon_i} + \abs{\epsilon_j} + \norm{ \bm{x}_i - \bm{x}_j }), \\
    & = \lambda^l(\lambda^f + 1) \norm{ \bm{x}_i - \bm{x}_j } + \lambda^l (\abs{\epsilon_i} + \abs{\epsilon_j}). 
    \label{eq:ResistModel:LossLipschitz}
\end{split}
\end{equation}

Denote the selected data samples as $(\bm{x}_j^c, y_j^c), \forall j \in \bm{s}$. 
Then assign each point $i$ of the entire dataset to a cluster centered by its nearest selected data sample $j$,  
and suppose that there are $k_j$ points within the cluster. 
Then the average loss of all data points is bounded as follows, 
\begin{equation}
\begin{split}
    \frac{1}{n}\sum_{i \in D} l_{\bm{w}_{\bm{s}}}(\bm{x}_i, y_i) 
        & = \frac{1}{n} \sum_{j \in \bm{s}} \sum_{i=1}^{k_j} l_{\bm{w}_{\bm{s}}}(\bm{x}_i, y_i), \\
        & \stackrel{(\UseMacro{Inequality3})}{\le} \frac{1}{n} \sum_{j \in \bm{s}} \sum_{i=1}^{k_j} \big( l_{\bm{w}_{\bm{s}}}(\bm{x}_j^c, y_j^c) + \abs{ l_{\bm{w}_{\bm{s}}}(\bm{x}_i, y_i) - l_{\bm{w}_{\bm{s}}}(\bm{x}_j^c, y_j^c) } \big), \\
        & \stackrel{(\UseMacro{Inequality4})}{\le} \frac{1}{n} \sum_{j \in \bm{s}} \sum_{i=1}^{k_j} \big( 0 + \abs{ l_{\bm{w}_{\bm{s}}}(\bm{x}_i, y_i) - l_{\bm{w}_{\bm{s}}}(\bm{x}_j^c, y_j^c) } \big), \\
        & \stackrel{(\UseMacro{Inequality5})}{\le} \frac{1}{n} \sum_{j \in \bm{s}} \sum_{i=1}^{k_j} 
        \big( \lambda^l(\lambda^f + 1) \norm{ \bm{x}_i - \bm{x}_j^c } + \lambda^l (\abs{\epsilon_i} + \abs{\epsilon_j^c}) \big), \\
        & = \frac{\lambda^l(\lambda^f + 1)}{n} \sum_{j \in \bm{s}} \sum_{i=1}^{k_j} \norm{ \bm{x}_i - \bm{x}_j^c } 
        + \frac{\lambda^l}{n} \sum_{j \in \bm{s}} \sum_{i=1}^{k_j}  (\abs{\epsilon_i} + \abs{\epsilon_j^c}), \\
        & = \frac{\lambda^l(\lambda^f + 1)}{n} \sum_{j \in \bm{s}} \sum_{i=1}^{k_j} \norm{ \bm{x}_i - \bm{x}_j^c } 
        + \frac{\lambda^l}{n} \sum_{i \in D} \abs{\epsilon_i} 
        + \frac{\lambda^l}{n} \sum_{i \in \bm{s}} k_i \abs{\epsilon_i^c}, \\ 
        & = \frac{\lambda^l(\lambda^f + 1)}{n} \sum_{j \in \bm{s}} \sum_{i=1}^{k_j} \norm{ \bm{x}_i - \bm{x}_j^c } 
        + \frac{\lambda^l}{n} \sum_{i \in D} \alpha_i | \epsilon_i |, \\
        & \quad \quad \textrm{where } \alpha_i = 
\begin{cases}
        1, & \quad i \in D \setminus \bm{s}, \\
        k_i+1, & \quad i \in \bm{s}, \\
\end{cases} \\
& \stackrel{(\UseMacro{Inequality6})}{\le} \frac{\lambda^l(\lambda^f + 1)}{n} \sum_{j \in \bm{s}} \sum_{i=1}^{k_j} \norm{ \bm{x}_i - \bm{x}_j^c }
+ \frac{\lambda^l}{n} \sum_{i \in D} \alpha_i \sum_{i \in D} \abs{ \epsilon_i },\\
& \stackrel{(\UseMacro{Inequality7})}{\le} \frac{\lambda^l(\lambda^f + 1)}{n} \sum_{j \in \bm{s}} \sum_{i=1}^{k_j} \norm{ \bm{x}_i - \bm{x}_j^c }
+ 2\lambda^l \sum_{i \in D} \abs{ \epsilon_i },\\
\end{split}
\label{eq:ResistModel:AvgLossBoundNoProb}
\end{equation}
Inequality (\UseMacro{Inequality3}) utilizes the fact that $a-b \le \norm{a-b}$. 
Inequality (\UseMacro{Inequality4}) uses the zero loss assumption. 
Inequality (\UseMacro{Inequality5}) embeds Eq.~\eqref{eq:ResistModel:LossLipschitz}. 
We assume $\epsilon_i$ and $\epsilon_j^c$ follow the same normal distribution, because they come from the same dataset. 
Inequality (\UseMacro{Inequality6}) leverages the fact that $\sum_i a_i b_i \le \sum_i a_i \sum_i b_i, \forall a_i, b_i \ge 0$. 
Inequality (\UseMacro{Inequality7}) cancels out $n$ by $\sum_{i \in D} \alpha_i = 2n$, 
where $|\epsilon_i|$ is a sample from an independent random half-normal distribution with mean $\sigma \sqrt{\frac{2}{\pi}}$ and variance $\sigma^2(1-\frac{2}{\pi})$.

\end{proof}
}